%% file: 0069-SIP-2014-PIEEE.tex
\documentclass[journal]{IEEEtran}
\usepackage{cite}
\usepackage{textcomp}
\usepackage{array}
\usepackage{subfigure}
\usepackage{graphicx}
\usepackage{amsmath}
\usepackage{amssymb}
\usepackage{mathptmx}
\usepackage{epsfig}
\usepackage{color}
\usepackage[normalem]{ulem}
\usepackage{wasysym}
\usepackage{fancyhdr}
\usepackage{multirow}

\begin{document}
%
\title{Bio-inspired Visual Motion Estimation}


\author{ \authorblockN{Garrick Orchard$^{\dagger}$ and Ralph Etienne-Cummings$^{\star}$ \textit{Fellow, IEEE}}

 \authorblockA{
    $^\dagger$Singapore Institute for Neurotechnology (SINAPSE), National University of Singapore, Singapore\\
    $^\star$Department of Electrical and Computer Engineering, Johns Hopkins University, Baltimore MD, USA\\
    Email: garrickorchard@nus.edu.sg, retienne@jhu.edu}
    }
\maketitle

\begin{abstract}
Visual motion estimation is a computationally intensive, but important task for sighted animals. Replicating the robustness and efficiency of biological visual motion estimation in artificial systems would significantly enhance the capabilities of future robotic agents. 25 years ago, in this very journal, Carver Mead outlined his argument for replicating biological processing in silicon circuits. His vision served as the foundation for the field of neuromorphic engineering, which has experienced a rapid growth in interest over recent years as the ideas and technologies mature. Replicating biological visual sensing was one of the first tasks attempted in the neuromorphic field. In this paper we focus specifically on the task of visual motion estimation. We describe the task itself, present the progression of works from the early first attempts through to the modern day state-of-the-art, and provide an outlook for future directions in the field.
\end{abstract}


\IEEEpeerreviewmaketitle

\section{Introduction}
Visual sensing is a computationally intensive, but crucial task for sighted animals \cite{lee1980optic}. Although motion estimation is just one aspect of visual sensing, its importance is easily understood when observing its wide range of uses in biology \cite{srinivasan2004visual}, including depth perception, ego-motion estimation, collision avoidance and triggering escape reflexes, time-to-contact estimation (landing control), prey detection and identification, segmentation by motion, and visual odometry.

The ability to reliably estimate visual motion in artificial systems would find applications ranging from surveillance and tracking, to visual flight control \cite{srinivasan2004visual}, to video compression, image stabilization, and even the computer mouse \cite{1Tanner}. However, the most relevant application of bio-inspired visual motion estimation is for embedded sensing onboard robotic agents capable of moving through and interacting with their environment, since this is precisely the function for which biological visual motion systems have evolved. Key characteristics which distinguish this application from others are: the sensor must operate in real-time, the sensor must operate under egomotion, the sensor must be physically carried by the agent, and the sensor should provide information which is relevant for enabling the agent to interact meaningfully with its environment. In fact, Wolpert argues in a recent TED talk that control of motion is the primary purpose of the brain \cite{WolpertTED}.

There are significant differences between biological and artificial systems regarding how visual information is acquired and processed. State-of-the-art modern visual motion estimation methods still rely on capturing sequences of images (frames) in rapid succession, even though the majority of data in these images is redundant \cite{StatisticsOfNaturalImages}. The problem of storing and transmitting this redundant information is partially overcome by using dedicated video compression ASICs, or in the case of standalone visual motion sensors, by computing on chip \cite{MousePatentAdan, MousePatentHartlove, MousePatentTanner, px4flowBlockMatching}. Nevertheless, these artificial approaches capture frames at pre-determined discrete time points regardless of the visual scene. On the other hand, biological retinae continuously capture data and perform a combination of compression and pre-processing in analog (using graded potentials) at the focal plane itself, with the visual scene largely driving when and where data is transmitted as spikes (voltage pulses) down the optic tract. Spikes are similar to digital pulses in artificial systems in that their signal amplitude can be restored and they are therefore particularly useful for communicating over longer distances, such as along the optical tract.

Processing also differs significantly between biological and artificial systems. Similarly to how artificial systems typically capture data at a constant rate, they must also compute at a constant rate to ensure all the captured data is processed, thus processing of visual information continues even if the scene is static. On the other hand, computation in biological systems is driven by the sparse captured data (spikes), in turn ensuring that neuron activation is sparse \cite{SparseCoding} (since neuron activation is driven by the sparse incoming data). This sparsity combined with the low power consumption of neurons which are not computing \cite{BOLDsignal} results in significant energy savings. Modern biologically inspired sensors generate sparse data (events) in response to activity in the scene \cite{DVStobi, ATIS} and this data can be used to drive sparse computation on modern neural simulator platforms \cite{NeuroGrid}. Together, these sensors and neural simulators allow both data capture and computation to scale with scene activity.

The architectures used for processing also differ. Typical artificial systems compute on a small number of parallel processors, each of which performs sequential operations in a precise repeatable manner, and operates at a timescale on the order of nanoseconds. On the other hand, biology relies on massively parallel processing using a very large number of imprecise computing elements (neurons), each of which operates on a timescale on the order of milliseconds \cite{TemporalPrecision}. However, parallelism in artificial systems is increasing, particularly for visual processing (GPUs), and emerging custom neural hardware platforms \cite{NeuroGrid, Spinnaker}. State of the art ASICs dedicated to visual motion estimation are also optimized to perform processing in as parallel a manner as possible.

Despite the imprecise nature of individual neurons, biological systems perform robustly and continue to do so even after the death of individual neurons. The same is not true of artificial systems, where a single fault can cause catastrophic failure of the entire system. Similar fault tolerance is highly desirable in artificial systems, especially as the number of transistors per device continues to increase, and as the size limits of silicon technology continue to be pushed. Even looking past silicon to nanotechnology, device yield continues to be a major challenge \cite{NanotechnologyYield}.

Beyond handling minor faults, biological systems are also able to learn and adapt to changes in the visual system itself \cite{OcularDominanceSuture}, as well as to different environments through visual experience \cite{EnvironmentEffect}, allowing them to operate effectively under a wide range visual of conditions. Such self-contained on-line learning and adaptation would also prove valuable for artificial systems, removing any need for manual tuning of parameters for operation in different environments.

Biology's robust processing, low power consumption, and ability to learn and adapt all present desirable characteristics for artificial systems and drive the field of bio-inspired sensing and computation.

In this paper we provide a brief introduction to the visual motion estimation problem and provide background on the methods used in traditional computer vision versus biological systems, before reviewing advances in bio-inspired visual motion estimation for artificial systems, presenting our own approach to the problem, and discussing future directions.

\section{The Visual Motion Estimation Problem}\label{sec:TheProblem}
When relative motion is present between an observer (eye or camera) and objects in a scene, the projections of these objects onto the image plane (retina or pixel sensor) will move. Visual motion estimation is the task of estimating how the projections of these objects move on the image plane.

\begin{figure}
\centering
  \includegraphics[width=0.5\textwidth]{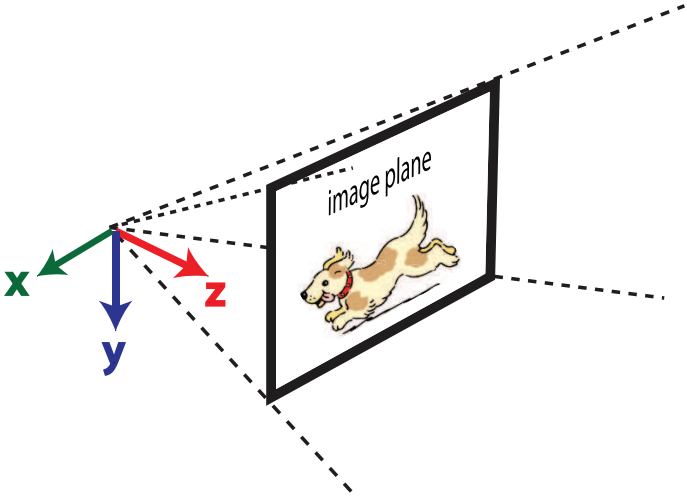}
    \caption{Definition of axes used in \eqref{eq:VisualMotion} for a pinhole camera approximation with unit focal length. The $z$ axis points out towards the scene perpendicular to the image plane, while the $x$ and $y$ axes are parallel to the image plane.}\label{fig:CameraSetup}
\end{figure}

Assuming a pinhole camera approximation, with the image plane at unit focal length, visual motion can be described as a function of the image plane co-ordinates $(x,y)$, the relative rotation $(\omega_x, \omega_y, \omega_z)$ and translation $(T_x, T_y, T_z)$ between the camera and the object being viewed, and the depth of the object $(z)$ \cite{jahne2000computer} as shown in Fig.~\ref{fig:CameraSetup} and \eqref{eq:VisualMotion} below.

\begin{equation}\label{eq:VisualMotion}
\begin{array}{l l l}
\frac{\delta x}{\delta t} &= &\frac{T_zx-T_x}{z} - \omega_y + \omega_zy + \omega_xxy - \omega_yx^2\\
\frac{\delta y}{\delta t} &= &\frac{T_zy-T_y}{z} + \omega_x - \omega_zx - \omega_yxy + \omega_xy^2\\
\end{array}
\end{equation}

Visual motion is constrained to lie in the image plane and therefore has no $z$-direction component. The first term in each equation describes the visual motion due to translation, which is depth dependent, while the remaining terms describe visual motion due to rotation, which is independent of depth. In other words, visual motion due to rotation does not depend on the structure of the scene, while visual motion due to translation does depend on scene structure. The rotations and translations in the equation above are for motion of the camera relative to the origin as depicted in Fig.~\ref{fig:CameraSetup}.

\begin{figure}
\centering
  \includegraphics[width=0.5\textwidth]{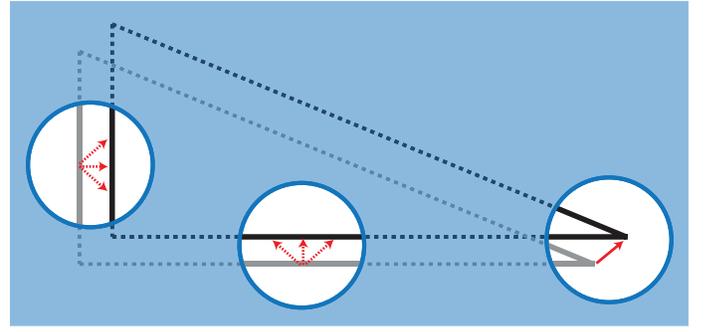}
    \caption{The aperture problem illustrated with a triangle moving between an initial position (grey) and a final position (black), while viewed only through three apertures (blue circles). For the leftmost aperture, the image only varies in the horizontal direction, and therefore only the horizontal component of motion can be estimated. Similarly, for the middle aperture, only the vertical component of motion can be estimated. For the rightmost aperture, the viewed image varies along both the horizontal and vertical directions and therefore motion can be uniquely determined.}\label{fig:ApertureProblem}
\end{figure}

The relationship described in \eqref{eq:VisualMotion} also shows that multiple different combinations of scene structure and relative motion can result in identical visual motion. Thus visual motion alone is not enough to infer relative motion or scene structure and additional information is required. For example, if the scene is static and the rotational motion of the sensor is known, then the translational motion direction can be determined, and a relationship between scene depth and camera translation speed can be obtained. Thus, measuring or even eliminating rotational motion allows additional valuable information to be derived from visual motion. 

Visual motion can only be estimated in the presence of an intensity gradient. A shape of uniform colour moving against a background of identical colour will have no intensity gradient and will therefore not elicit a visual motion stimulus. More specifically, for motion to be detected, the intensity gradient must be non-zero in the direction of motion. This is an example of the aperture problem \cite{aperture1988nakayama}, to which all visual systems are prone, and is illustrated in Fig.~\ref{fig:ApertureProblem}.

When considering only a small image region which has no intensity gradient in a particular direction, the magnitude of image velocity in that direction cannot be determined unless additional information is available. The component of motion in the direction of the maximum image gradient can be determined, and is known as the ``normal flow", since it is perpendicular (normal) to the edge orientation. The larger the image region under consideration, the more likely it will contain gradients in different directions, helping to alleviate the aperture problem. A common approach is to simultaneously consider multiple neighbouring image regions and assume their motion to be either consistent \cite{Lucas_Kanade_1981} or smoothly varying \cite{Horn81determiningoptical}, thus providing the additional constraint required to uniquely determine motion.

\section{Approaches to Visual Motion Estimation}\label{sec:Approaches}
For the purpose of providing background for later sections, we introduce here the basic theory underlying each of the three main classes of visual motion estimation approaches: correlation methods, gradient methods, and frequency methods.

Underlying all three of these methods is the assumption of brightness constancy, known as the brightness constancy constraint, which states that the brightness of a point remains constant after moving a small distance on the image plane $[\Delta x, \Delta y]$, within a small period of time, $\Delta t$. Formally this can be written as:
\begin{equation}\label{eq:ImageConstancy1}
\begin{array}{l l l}
I(x,y,t) &\approx & I(x+\Delta x, y+\Delta y, t+ \Delta t)\\
\end{array}
\end{equation}
where $I(x,y,t)$ is the intensity of the point located at $(x,y)$ on the image plane at time $t$.

\subsection{Correlation Methods}

Correlation methods for motion estimation rely on detecting the same visual feature at different points in time as it moves across the image plane. Correlation is used to determine whether two feature signals detected at different points in time relate to the same or different features. The feature signals on which correlation is computed can take the form of continuous-time analog signals, discrete-time analog signals, discrete digital signals, or even single bit binary tokens indicating only the presence or absence of a feature (``token methods"). The change in feature location can be combined with the change in time between detections to determine the feature's motion. The simplest features to use are brightness patterns, or derivatives thereof, the appearance of which remains constant over small time periods as described in \eqref{eq:ImageConstancy1}.

\begin{figure}
\centering
  \includegraphics[width=0.45\textwidth]{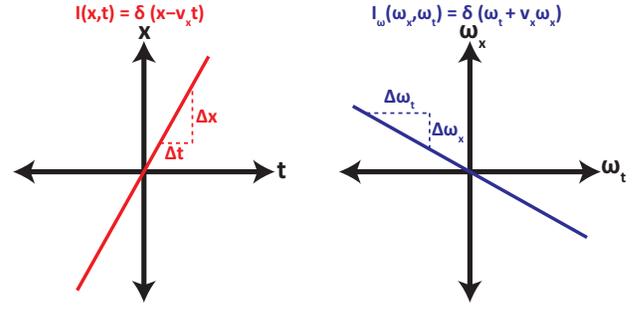}
    \caption{Frequency representation of the motion described in \eqref{eq:FrequencyMethod}. The left plot (red) shows the relationship between time and location for a point moving with constant velocity. The right plot (blue) shows the velocity dependent relationship between spatial and temporal frequency for the same moving point.}\label{fig:FourierFrequency}
\end{figure}

Most state of the art commercially available motion estimation ASICs rely on correlation methods \cite{MousePatentAdan, MousePatentHartlove, MousePatentTanner}. These devices capture frames at high frame rates and detect correlations in local pixel intensity patterns between images to determine motion. One very common approach is the Sum of Absolute Differences (SAD) block-matching algorithm \cite{px4flowBlockMatching}, which matches ``blocks" of pixels between frames by computing the SAD for pixel intensities. This process is repeated for many different blocks and the estimates from all of these blocks are combined to determine motion.

\subsection{Gradient Methods}
Gradient methods rely on the Taylor series expansion of \eqref{eq:ImageConstancy1}, which for a first order expansion can be rearranged into the form:
\begin{equation}\label{eq:ImageConstancy2}
\begin{array}{l l l}
\frac{\delta I(x,y,t)}{\delta x}\frac{\delta x}{\delta t} + \frac{\delta I(x,y,t)}{\delta y}\frac{\delta y}{\delta t} +\frac{\delta I(x,y,t)}{\delta t} &=& 0\\
\end{array}
\end{equation}
where $\frac{\delta x}{\delta t}$ and $\frac{\delta y}{\delta t}$ are the visual motion values that must be estimated, while  $\frac{\delta I(x,y,t)}{\delta x}$, $\frac{\delta I(x,y,t)}{\delta y}$ and $\frac{\delta I(x,y,t)}{\delta t}$ are intensity derivatives which can be obtained from captured frames.

Notice that if the intensity derivative in either spatial direction is zero, then motion in that direction is removed from the equation and cannot be estimated. This is the aperture problem discussed in Section~\ref{sec:TheProblem}. When these spatial derivatives are non-zero, but small, they are sensitive to noise and can still result in erroneous motion measurements. Even if accurate non-zero intensity derivatives are available, \eqref{eq:ImageConstancy2} is a single equation with two unknowns and thus does not provide a unique solution.

To arrive at a unique solution additional constraints must be imposed, such as that the motion of all points in an image patch will be equal (as is used in the Lucas-Kanade algorithm \cite{Lucas_Kanade_1981}), or that motion varies smoothly across image locations (as is used in the Horn-Schunk algorithm \cite{Horn81determiningoptical}).

\subsection{Frequency Methods}
Frequency based methods rely on the observation that there is a relationship between temporal frequency, spatial frequency, and velocity \cite{Adelson:85}. For simplicity consider a point (Dirac Delta function \cite{dirac1981principles}) moving in the $x$ direction. This point will trace out a line in the space-time plot on the left of Fig.~\ref{fig:FourierFrequency} with slope equal to the velocity. Taking the Fourier transform results in a line in frequency space with slope equal to the inverse of velocity, as shown on the right of Fig.~\ref{fig:FourierFrequency}.

\begin{equation}\label{eq:FrequencyMethod}
\begin{array}{l l l}
I(x,t) &=& \delta(x-v_x t)\\
v_x &=& \frac{\Delta x}{\Delta t}\\
I_{\omega}(\omega_x,\omega_t) &=&  \delta(\omega_t+v_x\omega_x)\\
\frac{1}{v_x} &=& -\frac{\Delta\omega_x}{\Delta\omega_t}\\
\end{array}
\end{equation}
where $I(x,t)$ is the intensity at location $x$ at time $t$, $v_x$ is the velocity in the $x$ direction, $\delta$ is the Dirac Delta function, and $I_{\omega}(\omega_x,\omega_t)$ is the Fourier transform \cite{Fourier} of $I(x,t)$.

The case described above is an ideal case where the stimulus is a point (Dirac Delta function) and therefore has equal energy at all spatial frequencies. In the more general case of a stimulus with an arbitrary distribution of spatial frequency content, the energy will still be constrained to lie along the line shown on the right of Fig.~\ref{fig:FourierFrequency}.

\begin{figure}
\centering
  \includegraphics[width=0.45\textwidth]{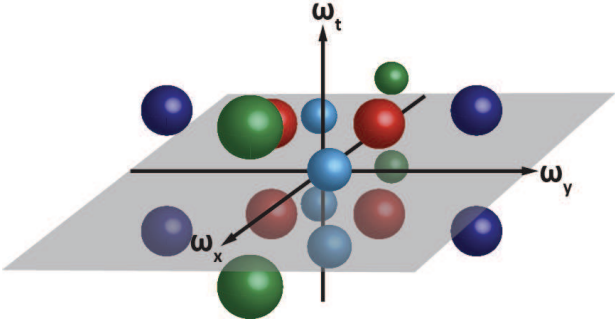}
    \caption{An example placement of four different quadrature pairs of spatiotemporal filters in frequency space. Each quadrature pair is indicated by a different color and is symmetric about the $\omega_t$ axis and the plane $\omega_t=0$ (shaded gray). The red and dark blue quadrature pairs are sensitive to different velocities in the $y$-direction. The green quadrature pair is sensitive to motion of a specific speed in the $x$-direction. The light blue quadrature pair is most sensitive to a particular velocity which has both $x$ and $y$ components.}\label{fig:FilterPlacement}
\end{figure}

Visual motion can be estimated by finding the slope of the line in Fig.~\ref{fig:FourierFrequency}, which can be achieved by tiling frequency space with spatiotemporal filters as shown in Fig.~\ref{fig:FilterPlacement} and combining their responses to find the location of the energy peak.

\subsection{Implementation}
Typical artificial approaches to motion estimation rely on a frame-based camera to capture snapshots (frames) of the scene at fixed intervals. A processor is then used to apply one of the three methods described above to estimate visual motion. In computer vision typically correlation or gradient based methods are used, with frequency methods regarded as bio-inspired approach as will be discussed in Section~\ref{sec:BioMotion}.

In all three of the methods outlined above, increasing the frame rate improves the accuracy of the algorithm because the brightness constancy constraint relies on the assumption of a small time period between observations (frames).

For correlation methods, increased frame-rate also decreases the distance a feature can move between frames, thereby helping to restrict the search region for that feature in subsequent frames. For example, typical optical mouse algorithms operate at thousands of frames per second (albeit small frames), and the search can be in increments of fractions of a pixel. For gradient and frequency methods, increased frame-rate is equivalent to a higher temporal sampling rate, reducing aliasing and allowing for higher order digital filters to be used when estimating the temporal derivative or frequency.

However, increasing frame-rate also increases the computing power required to sustain real-time operation, since more frames must be processed within the same time period. The additional computing needs are typically met by using more powerful hardware, such as GPUs, FPGAs, and custom ASICs. Tight coupling of a frame-based sensor and ASIC is often used to reduce communication costs for embedded applications, such as for a stand alone motion estimation unit relying on a high frame rate, or in-camera video compression relying on block matching.

Some artificial approaches to motion estimation do not rely on frames, but instead process on continuous-time analog signals derived from CMOS photodiodes. Notable examples include \cite{1Tanner, 39AlanStockerMain}.

For all three of the approaches described, motion estimates within a local image region can be computed independently of motion estimates for other image regions. This opens up the possibility of simultaneously computing motion for different image regions in parallel. GPUs, FPGAs, and ASICs all take advantage of this.

The computational complexity of each approach varies. A major disadvantage of gradient methods is that they typically require the expensive computation of a matrix inverse (or pseudo-inverse) in order to find the solution which best satisfies (with the least square error) the brightness constancy \eqref{eq:ImageConstancy2} and secondary constraints. However, gradient methods have the advantage of computing on instantaneous gradient values and therefore require very little memory (only enough to estimate temporal gradients).

Computing correlations for correlation methods is computationally simpler, but requires the system to have memory of previously observed features, whether by delaying feature signals or by explicitly storing them.

Digital implementations of frequency methods require even more memory because many time points are required to detect temporal frequencies, particularly if very low frequencies are present. The spatial and temporal frequency content of the scene is typically not known in advance, so frequency methods require a large number of spatiotemporal filters in order to accurately detect the frequency content of different motion stimuli. Implementing these filters digitally is costly both in terms of computation and memory.

The robustness of the approaches also varies. When computing the correlation between two signals, it is not always clear whether a large output has resulted from a strong correlation or from large input signals. Signals can be normalized before correlation to overcome this ambiguity. However, the typical signals on which correlation is computed arise from a combination of image motion and image spatial contrast and it is not always possible to disambiguate the effects of motion versus contrast in the final output.

Gradient methods rely on the ratio of the temporal gradient to the spatial gradient \eqref{eq:ImageConstancy2} and are therefore very sensitive to noise, particularly when spatial gradient signals are weak.

Spatiotemporal frequency models are far more robust to noise, but as discussed above, the computational and memory requirements for estimating frequencies is far higher.

\section{Visual Motion Estimation in Biology}\label{sec:BioMotion}

\begin{figure*}
\centering
  \includegraphics[width=0.9\textwidth]{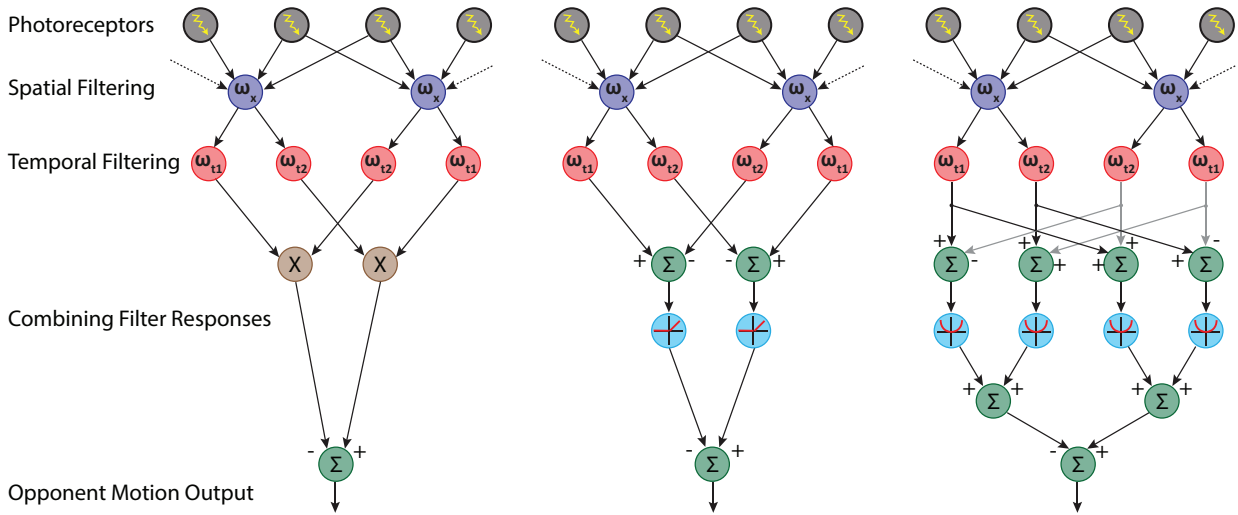}
    \caption{A comparison of the Hassenstein-Reichardt (left), Barlow-Levick (center), and Adelson-Bergen (right) models, similar to that in \cite{32Higgins}. Spatial filters (dark blue) can either differ in location (as shown) or in phase. Two temporal filters (red) are used, with the second ($\omega_{t2}$) having a longer delay than the first ($\omega_{t1}$). The Reichardt model detects correlation between the signal at one location and the delayed signal from a neighbouring location. The correlation can be modelled as a multiplication (brown), or as a logical AND if using single bit signals. The Barlow-Levick model instead uses the signal to block response to motion in the null direction. This can be modelled as a summation (green) followed by half wave rectification (light blue) to prevent negative intermediate responses. If using 1-bit signals, the inhibition can be modelled as an AND gate with the inhibiting (delayed) input negated. The Adelson-Bergen model combines separable spatiotemporal filter responses incorporating a squaring (light blue) non-linearity to compute motion energy. The output of the Adelson-Bergen model is formally equivalent to that of the Hassenstein-Reichardt model \cite{Adelson:85}.}\label{fig:MotionModels}
\end{figure*}

The animal kingdom is incredibly diverse and the visual systems of many creatures have evolved independently \cite{EyeEvolution} (although they may share a common origin), resulting in variations in size, number, shape, location, wavelength sensitivity, and acuity of eyes across families \cite{AnimalEyes}. Similarly to how eyes vary by family, so does the process of visual motion estimation. For the sake of discussion we will focus specifically on \textit{Drosophila} (fruit flies) and macaque (monkeys), which are both well studied genera, but possess very different visual systems. Properties of the vision systems of \textit{Drosophila} and macaque generalise to many other insects and mammals respectively. This section is intended only as a brief introduction. For more details, the reader is directed to neuroscience reviews covering \textit{Drosophila} \cite{borst2010fly} and primate vision \cite{martin2011information}.

Despite the differences between macaque and \textit{Drosophila} vision systems, there are also many similarities. In both systems initial computation is performed at the focal plane by neurons which communicate using a combination of spiking and graded responses. These neurons respond to intensity changes, with responses to intensity increases (ON) and decreases (OFF) processed independently by parallel pathways \cite{FunctionalSpecialization}. In \textit{Drosophila}, motion estimation can still be performed in the L1 (ON) pathway if the L2 (OFF) pathway is blocked, and vice versa. 
In macaque, direction selective Starburst Amacrine Cells (SACs) \cite{RoleOfStarburstAmacrine} 
can be found in the retina itself, and separate SACs are used to process ON and OFF responses in parallel. Although direction selective, these cells show very limited speed sensitivity, with true velocity sensitive neurons only found in higher visual areas.

In \textit{Drosophila}, as with many other animals, a fast escape reflex triggered by visual motion aids in evading approaching predators. Low latency motion detection is critical for this task and is achieved by keeping the motion processing circuitry relatively simple and located close to the photoreceptors \cite{drosophilaVision}. These motion processing circuits rely on correlation methods, which are fast, and allow for compact implementation which can be realised in the limited space available near the photoreceptors \cite{reichardt1961autocorrelation}. Furthermore, these circuits are tuned to detect stimuli characteristics indicative of an approaching predator, rather than to accurately measure a wide range of complex motion stimuli.

On the other hand, the macaque visual system is more concerned with accurately estimating motion for a wide range speeds and visual stimuli than detecting approaching predators. Accurate motion estimates help to achieve a deeper scene understanding, which can then be used for action planning. This link between motion estimation and scene understanding requires interaction between motion detectors and cortex, and motion sensitive neurons are therefore found in various cortical areas \cite{CortexMotion}.

Low latency detection is also desirable for macaque, but is not as important as when triggering escape reflexes, allowing the luxury of using more complex motion estimation methods and taking advantage of the computational resources available in cortex. The macaque visual system is therefore not restricted to using simple correlation methods. The presence of spatiotemporal frequency sensitive neurons in the Middle Temporal (MT) visual area, which plays an important role in motion estimation \cite{MotionMT}, suggests that motion estimation in primates relies on frequency methods \cite{SeeingThings}.

In both macaque and \textit{Drosophila} the visual motion estimation system is tightly coupled with the motor system. In macaque, the vestibular ocular reflex is important for visual perception \cite{VestibularOcularReflex}. Visual inputs can trigger saccades, and saccades can suppress visual responses \cite{SaccadeSuppression}. In \textit{Drosophila}, the visual system can trigger motor responses through the optomotor reflex, and flight control is heavily reliant on visual motion estimation \cite{srinivasan2004visual}. 

It is therefore important when considering biological vision systems to note that they do not exist in isolation. The visual system is part of an embodied system capable of moving through, and interacting with the environment. The visual and motor systems are tightly coupled and deficits in either system can affect the other, as documented in both primates \cite{HumanVisualMotor} and \textit{Drosophila} \cite{optomotorFly}. The optomotor response is so strong in many insects that motor outputs in response to visual stimuli can provide insight into the visual system .

It was through an investigation of the optomotor response of the beetle Clorophanus in the 1950s that Hassenstein and Reichardt arrived at their seminal model of the Elementary Motion Detector (EMD) \cite{reichardt1961autocorrelation}, shown in Fig.~\ref{fig:MotionModels}a.
The Hassenstein-Reichardt EMD computes the correlation between the signal of one photoreceptor and the time-delayed signal of a neighbouring photoreceptor. The delay is typically modelled as a low pass filter and correlation performed as multiplication. Strong correlation indicates the presence of motion in the preferred direction. A mirror symmetric circuit detects motion in the opposite direction, and the difference between the circuit outputs indicates motion direction.

The Hassenstein-Reichardt EMD does not provide a direct measure of the stimulus velocity. Instead, it provides an indication of how well the motion stimulus matches the EMD's preferred combination of speed and spatial frequency. Multiple combinations of speed and spatial frequency can result in the same EMD output magnitude, so speed cannot be uniquely determined. Even if the spatial frequency is known, there are speeds greater than and less than the EMD's preferred speed for which the response magnitude would be equal, so speed would still not be uniquely determined.

These observations led to many interesting predictions which were later verified. Evidence of the existence of the Hassenstein-Reichardt model has since been found in many other visual systems, including that of \textit{Drosophila}.

By the 1960s Hubel and Weisel had isolated directionally selective units in the cat cortex \cite{HUBELWeisel}. Later similar responses were observed in the tectum of pigeons and frogs.
Barlow and Levick found such directional responses even earlier in the visual pathway of the rabbit, in the retina itself. Based on their recordings, they proposed what is now known as the Barlow-Levick model \cite{barlow1965mechanism}
shown in Fig.~\ref{fig:MotionModels}b.
The Barlow-Levick model relies on inhibition instead of excitation as the underlying mechanism, with null direction motion inhibiting a motion unit's response. Although the presence of Hassenstein-Reichardt and Barlow-Levick motion models have been ruled out in macaque,
a similar mechanism relying on unbalanced inhibition is thought to underly directional selectivity of starburst amacrine cells in primate retina \cite{RoleOfStarburstAmacrine}.

As mentioned earlier, the presence of cells in MT sensitive to specific spatiotemporal frequencies indicates that frequency methods likely underly motion perception in macaque. In 1985, Adelson~and~Bergen \cite{Adelson:85} and Watson~and~Ahumada \cite{Watson:85} proposed similar architectures for motion estimation based on spatiotemporal frequency filtering. The Adelson-Bergen motion energy model is shown in Fig.~\ref{fig:MotionModels}c. Each unit computes the energy at a specific spatiotemporal frequency, with the relationship between spatial and temporal frequency being indicative of speed as outlined in Section~\ref{sec:Approaches}. Adelson and Bergen also showed how the opponent energy output by their model was equivalent to the output of the Hassenstein-Reichardt model.
The Adelson-Bergen model has since been used to explain many visual illusions \cite{Fermuller2010315, illusionsweiss2002motion} and Simoncelli and Heeger \cite{SimoncelliMT} have proposed a model describing how the spatiotemporal responses in MT may be computed in biology.

Simoncelli \cite{simoncelli2003local} also took the brightness constancy constraint \eqref{eq:ImageConstancy2} relied upon by gradient methods and used it to develop a probabilistic Bayesian framework capable of explaining responses of MT neurons, although the framework does not describe how such responses are computed physiologically.

In the abovementioned models, individual motion units do not encode velocity directly, but rather are selective to specific spatiotemporal stimuli. To infer velocity, the responses of motion units must be combined as described in \cite{35EtienneCummings}.

\section{Review of Bio-inspired Works}

\begin{table*}
\caption{Summary of bio-inspired VLSI visual motion estimation works}
\label{table:Summary}
\centering
\begin{tabular}{|l|c|r|l c|l|l|l|}
\hline
\multicolumn{1}{|c|}{Author}                  & \multicolumn{1}{|c|}{Year}  & \multicolumn{1}{|c|}{Process} & \multicolumn{2}{|c|}{Array}  & \multicolumn{1}{|c|}{Motion} & \multicolumn{1}{|c|}{Method}          & \multicolumn{1}{|c|}{Feature}                  \\ \hline \hline
    Tanner \cite{1Tanner}            & 1986 & 2$\mu$     & 1D & (1x16) & Global & Block-Matching  & Intensity variation      \\ \hline
    Tanner  \cite{1Tanner}           & 1986 & 1.5$\mu$   & 2D &(8x8)  & Global & Gradient        & -     \\ \hline
    Franceschini \cite{Franceschini} & 1989 & - & 1D & (1x100) & Local   & Token & Temporal edge \\ \hline
    Andreou \cite{3Andreou}          & 1991 & 2$\mu$     & 1D &(1x25) & Global & Reichardt             & ON-center OFF-surround     \\ \hline
    Etienne-Cummings \cite{6EtienneCummings}  & 1992 & 2$\mu$  & 2D &(5x5)  & Global  & Token (TI)      & Temporal edge of center surround  \\ \hline
    Horiuchi  \cite{4Horiuchi, 7Horiuchi}       & 1992 & 2$\mu$     & 1D &(1x17) & Local  & Token (TS)          & Temporal edge       \\ \hline      Delbruck   \cite{8Delbruck}        & 1993 & 2$\mu$     & 2D &(25x25) & Local & Reichardt     & Temporal contrast    \\ \hline
    Sarpeshkar \cite{43Sarpeshkar, 44Sarpeshkar}   & 1993 & 2$\mu$     & NA & NA&NA      & Token         & Temporal edge of spatial contrast\\ \hline
    Gottardi  \cite{40Gottardi}        & 1995 & 2$\mu$     & 1D &(1x115) & Global & Block-Matching  &Intensity values  \\ \hline
    Kramer \cite{10Kramer, 11Kramer}      & 1995 & 2$\mu$     & 1D &(1x8)  & Local & Token (FS)      &Temporal edge     \\ \hline
    Yakovleff     \cite{19Yakovleff}        & 1996 & 2$\mu$  &  1D &(1x61) & Local  & Block-Matching & Sign of spatiotemporal gradients \\ \hline         Arreguit \cite{Andrearreguit1996cmos} & 1996 & 2$\mu$  &  2D &(\~9x9) & Local & Block-Matching & Spatial edge \\ \hline
    Etienne-Cummings \cite{14EtienneCummings, 15EtienneCummings}  & 1997 & 2$\mu$  &  2D &(9x9) & Global            & Token & Temporal edge of center surround\\ \hline 
    Moini   \cite{16Moini, moini2000vision}      & 1997 & 1.2$\mu$  & 2x1D &(2x64) & Local  & Block-Matching            & Spatiotemporal templates      \\ \hline
    Harrison  \cite{18Harrison}          & 1998 & 2$\mu$   & 2D &(1x2)  & Local  & Reichardt       & Temporal contrast   \\ \hline
    Higgins      \cite{23Higgins}     & 1999 & 1.2$\mu$ & 2D &(14x13) & Local  & Token (ITI, FS)   & Temporal edge    \\ \hline
    Indiveri     \cite{20Indiveri}     & 1999 & 1.2$\mu$ & 2D &(8x8)    & Global   & Token (FS)    & Temporal edge    \\ \hline
    Jiang        \cite{21Jiang}     & 1999 & 0.6$\mu$ & 2D &(32x32)    & Global & Token (ISI)    & Temporal edge of spatial contrast  \\ \hline
    Etienne-Cummings  \cite{35EtienneCummings}  & 1999 & 2$\mu$ & 2x1D &(2x18)   & Global   & Adelson-Bergen & Spatiotemporal energy of edge map               \\ \hline
    Barrows \cite{47Barrows}     & 2000 & 1.2$\mu$ & 2x1D & (2x4)     & Global & Token (FS) &  Spatial features                                  \\ \hline
    Liu \cite{24Liu}     & 2000 & 1.2$\mu$ & 1D &(1x37)     & Global  & Reichardt    &  Temporal contrast                                  \\ \hline
    Pant       \cite{31Pant}       & 2000 & 1.6$\mu$   & 2D &(13x6)    & Local & Reichardt   & Temporal contrast                                     \\ \hline
    Higgins    \cite{25Higgins}       & 2000 & 1.2$\mu$   & 2D &(13x15)  & Local  & Token (FS)  & Temporal edge                                         \\ \hline
    Harrison  \cite{27Harrison}        & 2000 & 1.2$\mu$   & 1D &(1x22)   & Global    & Reichardt         & Temporal contrast                          \\ \hline
    Higgins   \cite{29Higgins}       & 2002 & 1.2$\mu$ & 2D &(27x29)      & Local     & Token (ITI)        & Temporal edge                         \\ \hline
    Yamada  \cite{28Yamada}     & 2003  & 1.5$\mu$ & 2D &(2x10)       & Local     & Token (FS)    & Spatial edge \\ \hline
    Ozalevi     \cite{30Ozalevli}     & 2003   & 1.5$\mu$ & 2D &(6x6)       & Global    & Adelson-Bergen & Spatiotemporal energy       \\ \hline        Massie      \cite{50Massie}     & 2003   & 0.5$\mu$ & 12x1D &(12x90)     & Yaw/Pitch/Roll    & Token (FS) &  Temporal edge of spatial contrast\\ \hline
    Stocker     \cite{39AlanStockerMain}    & 2004 & 0.8$\mu$  & 2D &(30x30)    & Local & Gradient      & -       \\ \hline
    Ozalevi     \cite{33Ozalevli, 34Ozalevli}  & 2005 & 1.6$\mu$ & 2D &(6x7)     & Global       & Reichardt      & Temporal contrast       \\ \hline        Harrison  \cite{51Harrison}        & 2005 & 0.5$\mu$ & 2D &(16x16)   & Global       & Reichardt      & Temporal contrast       \\ \hline
    Shoemaker  \cite{97shoemaker2005insectVLSI} & 2005 &  0.35$\mu$ & 1D  & (1x7)  & Global & Reichardt  & Temporal contrast \\ \hline Mehta     \cite{Mehta2}            & 2006 & 0.5$\mu$ & 2D &(95x52)   & Local       & Gradient      & -       \\ \hline
    Moeckel   \cite{52Moeckel}       & 2007 & 1.5$\mu$ & 1D &(1x24)    & Local & Token (FS) & Temporal contrast       \\ \hline
    Bartolozzi  \cite{36Bartolozzi}  & 2011 &  0.6$\mu$ & 1D &(1x64)   & Local     & Token (FS)     & Temporal edge \\ \hline
    Roubieu \cite{99roubieu2013two} & 2013 & - & 1D   & (1x5)         & Global    & Token (ISI) & Temporal contrast \\ \hline
\end{tabular}
\end{table*}

In this section we discuss integrated real-time bio-inspired visual motion estimation works, which are compared in Table~\ref{table:Summary}.
The discussion is divided into three subsections, one for each of the three methods described in Section~\ref{sec:Approaches}, and within each subsection, works are described in chronological order.

\subsection{Gradient Methods}
In his seminal 1986 thesis, Tanner \cite{1Tanner} presented two VLSI implementations of visual motion estimation. The first made use of the correlation method and will be discussed in the next subsection. Tanner's second implementation used a 2D analog VLSI gradient based approach relying a feedback loop to arrive at a minimum-error solution which simultaneously satisfies both the brightness constancy constraint \eqref{eq:ImageConstancy2} and a local smoothness constraint. This gradient based approach was later adopted and extended in other works by Stocker \cite{AlanBook} and Mehta \cite{Mehta2}.

\subsection{Correlation Methods}

\subsubsection{Block Matching}
Tanner's second implementation uses a correlation method. It comprises a linear array of 16 pixels which is sampled and binarized in a single step to produce a binary image. These binary images are captured and their correlation to shifted versions of an initial image are computed (the shifts are 1 pixel left, no shift, and 1 pixel right). Digital pulses indicate when leftward or rightward displacement by a single pixel has occurred, at which point a new initial image is captured and the process is repeated. Velocity is encoded by the digital pulse rate at the output of the sensor.
Gottardi \cite{40Gottardi} would later present a similar approach using CCD pixels coupled with CMOS circuity, but capable of detecting motion of up to 5 pixels per image.

Yakovleff \cite{19Yakovleff} presented a local matching approach which uses the sign of temporal gradients as the feature to be matched by digital circuits.

Arreguit \cite{Andrearreguit1996cmos} presented the first 2D array for block matching, which used spatial edges as features and was designed for use as a pointing device (computer mouse). Pixels computed motion locally, and local estimates were combined to estimate global motion.

\subsubsection{Hassenstein-Reichardt and Barlow-Levick Models}
At a high level, Tanner's implementation can be seen as sequentially implementing Reichardt detectors with increasing time delays until sufficient correlation is detected. In 1991 Andreou \cite{3Andreou} reported an analog implementation of the Reichardt detector which instead outputs the correlation value itself. The sensor computes ON-center OFF-surround features in analog and uses an all-pass filter to implement the delay. Outputs from all the Reichardt detectors along a linear array are summed and output as a differential current.

Delbruck \cite{8Delbruck} later presented a 2D analog variation of the Reichardt detector. Temporal contrast was used as the input signal and was delayed using a multi-tap analog delay line capable of propagating a signal across multiple pixels. At each pixel the correlation is computed between the propagating signal and the local signal before being propagated to the next pixel. The output therefore incorporates signals from many pixels and increases if motion is sustained across multiple pixels. However, the magnitude of the output is highly dependent on contrast.

Harrison \cite{27Harrison} presented another analog Reichardt implementation which also computes correlation based on temporal contrast, but exhibits decreased dependence on contrast magnitude. The detectors in the chip are either accessed individually, or combined using a linear summation to obtain a global response. Harrison also used the approach to generate an artificial optomotor response (torque signal) and compared it to measurement of \textit{Drosophila} in experiments \cite{27Harrison}.

Pant \cite{31Pant} presented an analog Reichardt implementation in which the responses of multiple Reichardt detectors are combined on-chip in a non-linear fashion. Pant showed how the output can be used to generate a torque signal to control the gaze direction of a robot during visual tracking, even though adjusting the gaze of the robot induces optical flow through egomotion \eqref{eq:VisualMotion}.

Liu \cite{24Liu} also presented an analog Reichardt implementation with non-linear on-chip integration of detector outputs, and showed that the frequency response of the chip is similar to the frequency response of Horizontal System (HS) neurons in \textit{Drosophila} \cite{optomotorFly}.

\begin{figure}
\centering
  \includegraphics[width=0.5\textwidth]{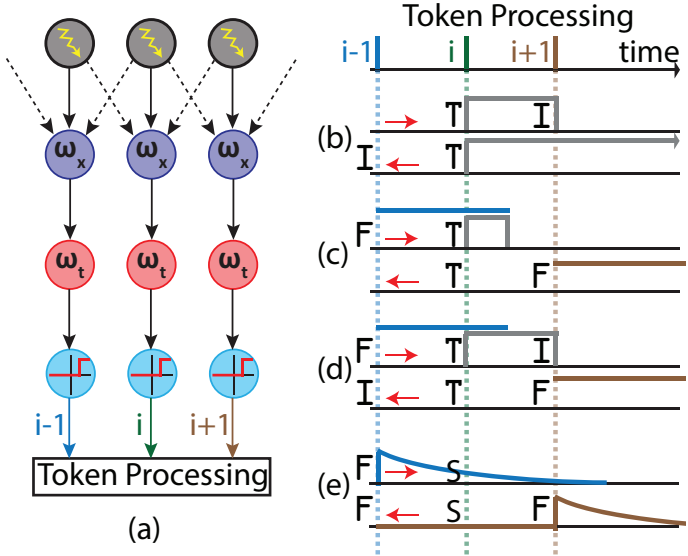}
    \caption{Popular token methods. The left image (a) shows how tokens are typically generated by thresholding (light blue) the output of a temporal bandpass filter (red) to detect changes in pixel intensities. Spatial filtering (blue) is optionally performed as a first step. The Trigger and Inhibit (TI), Facilitate and Trigger (FT), Facilitate Trigger and Inhibit (FTI), and Facilitate and Sample (FS) methods are shown in (b-e) respectively for a rightward moving stimulus, which will trigger responses from left (first) to right (last) when moving across the array. Red arrows indicate preferred direction of motion in each case.}\label{fig:TokenMethods}
\end{figure}

Shoemaker \cite{97shoemaker2005insectVLSI} presented an analog Reichardt implementation based on visual motion processing in the fly. The sensor makes use of non-linearities in the processing chain to reduce the dependence of the sensor output on scene contrast and spatial frequencies.

Harrison \cite{51Harrison} later presented a time-to-contact sensor for collision avoidance based on detecting 2D looming motion fields with Reichardt detectors.

Although partially addressed in the latter of the works mentioned thus far, one of the shortcomings of the Reichardt detector is the dependence of the output on stimulus contrast. This can be overcome by using so-called ``token" methods, where thresholding provides a 1 bit token to signal the presence or absence of a stimulus.

\subsubsection{Token Methods}
Horiuchi \cite{4Horiuchi} developed a linear array which used detection of a sufficient temporal intensity derivative (edge) as a digital token. Tokens from neighbouring pixels propagate down a digital delay line in opposite directions until they cross. The location in the delay lines at which they cross indicates both speed and direction. The delay line between each pixel pair contributes a ``vote" into a Winner-Take-All 
circuit which outputs the location with the most votes.

Etienne-Cummings \cite{6EtienneCummings} developed a chip for 2D motion detection which uses the temporal derivative of a center-surround feature as the token. The chip encoded speed using the width of a pulse initiated by departure of a token from one pixel and terminated by arrival of the token at either neighbour. A voting scheme was used to determine direction.

Sarpeshkar \cite{43Sarpeshkar} also took a token approach in which tokens trigger digital pulses. The pulse from one pixel would be delayed and correlated against its neighbour. The output provided a measure of how well the observed stimulus matched the circuit's optimal stimulus, but the response to non-optimal is ambiguous because both increasing and decreasing the stimulus speed causes the response to decrease. In the same paper, Sarpeshkar proposed a facilitate and trigger approach to overcome the speed ambiguity.

Similar approaches to that proposed by Sarpeshkar soon became popular. Fig.~\ref{fig:TokenMethods} outlines some of these token methods. The Trigger and Inhibit (TI) mechanism (Fig.~\ref{fig:TokenMethods}b) triggers a pulse when a token is detected at a pixel, and ends (inhibits) the pulse when the token is detected at the next pixel, thereby providing a pulse with width inversely proportional to speed. However, for motion in the null direction, the inhibition occurs before the trigger, and the pulse can continue indefinitely. The shortest of two pulses from directionally opposing circuits is typically assumed to be the correct one.

The Facilitate and Trigger (FT) method (Fig.~\ref{fig:TokenMethods}c) only triggers a pulse if a facilitation signal generated by the previous pixel is present. The pulse ends when the facilitation signal ends, thereby limiting the maximum pulse width to the width of the facilitation pulse. For motion in the null direction, the trigger occurs before the facilitation signal and no pulse is generated. Unlike the TI case, the pulse from the FT approach is directly proportional to speed.

The Facilitate, Trigger, and Inhibit (FTI) method (Fig.~\ref{fig:TokenMethods}d) combines the outputs of three pixels, using a facilitation signal for directional selectivity, followed by a Trigger and Inhibit mechanism to generate a pulse inversely proportional to the stimulus speed.

In the TI, FT, and FTI methods (Fig.~\ref{fig:TokenMethods}b-d) the duration a pulse must be measured in order to infer speed. The Facilitate and Sample (FS) method (Fig.~\ref{fig:TokenMethods}e) overcomes this by using a shaped facilitation pulse. Instead of triggering an output pulse, the facilitation pulse is sampled providing a value proportional to speed. For motion in the null direction, the facilitation signal will be zero when it is sampled.

Other methods have been proposed which take a similar approach, but rely on inhibition to suppress response in the null direction rather than facilitation to enable response in the preferred direction.

Kramer \cite{10Kramer, 11Kramer} presented the first FTI and the first FS token implementations, each using an 8 pixel linear array with temporal contrast edges as the token.

Etienne-Cummings \cite{14EtienneCummings} developed a foveated sensor for tracking and stabilization, consisting of a 19x17 pixel array for detecting onset and offset of spatial edges, with the middle 5x5 pixels replaced by a 9x9 array of smaller motion estimating pixels which output motion direction only, thereby realising a ``bang-bang" output.

Higgins \cite{23Higgins} later presented the first Inhibit, Trigger, and Inhibit (ITI) implementation as well as the first 2D FS implementation, both using temporal contrast edges for the token. His FS implementation also subtracted the samples of opposite direction circuits on-chip to provide a signed velocity. He later further developed the concept and extended it to larger array \cite{29Higgins}. Jiang \cite{21Jiang} meanwhile presented the first 2D FTI implementation, using temporal edges of spatial contrast as the token, while Yamada \cite{28Yamada} demonstrated an FS token implementation using 1D arrays and reported on its possible application to traffic flow measurement and monitoring blind corners while driving.

As neuromorphic front end sensors for temporal contrast detection improved, Higgins \cite{25Higgins} and Indiveri \cite{20Indiveri} both adopted multi-chip approaches to motion-estimation, relying on a stand-alone specialized front end sensor for temporal contrast detection, and a separate chip for the FS token algorithm implementation. The multi-chip approach carries the disadvantage of requiring additional power for off-chip communication between the front-end sensor and the motion computation chip. However, moving the in-pixel motion computation circuits to a separate chip reduces pixel size on the front-end chip, allowing a denser pixel array at the front-end.

Barrows \cite{47Barrows} implemented a multi-chip token method which combined a microcontroller for postprocessing with a front end sensor for extracting programmable spatial features. Different spatial features were found to work well for different stimuli, and the combined use of multiple spatial features was used to improve performance. The chips were used to control the rudder of a Micro Aerial Vehicle (MAV) to help it avoid obstacles.

The multi-chip approach also allows signals to be remapped between the front-end sensor and back-end motion computation chip, allowing mapping from cartesian to polar co-ordinates, which can be useful for measuring looming motion fields as Higgins and Indiveri both demonstrated \cite{25Higgins, 20Indiveri}. Furthermore, signals can easily be copied and routed to multiple motion processing chips. Higgins demonstrated such an approach for simultaneously computing motion in cartesian and polar co-ordinates using two motion processing chips in parallel. Higgins also demonstrated how two front-end sensors can feed into a single motion processing chip to compute motion only at a specific disparity (depth in $z$ direction) \cite{25Higgins}.

Massie \cite{50Massie} presented a combination imager and motion estimation chip for roll, pitch, and yaw estimation. The chip consists of 12 linear 90 pixel arrays (2 for yaw, 2 for pitch, and 8 for roll) relying on the token based FS method. Integrated into the same chip was a 128x128 pixel variable acuity imager capable of providing maximum resolution on objects of interest, while conserving bandwidth by combining pixel responses from ``uninteresting" regions.

Ozalevi \cite{33Ozalevli} presented a multi-chip approach which used a separate front end sensor to generate temporal edge tokens, but a low-pass filter was used to convert these tokens back into analog signals which were processed by separate chips implementing Hassenstein-Reichardt and Barlow-Levick models. The low pass filter also serves to create the delays required by these models (see Fig.~\ref{fig:MotionModels}). Thus, an analog implementation of the Hassenstein-Reichardt and Barlow-Levick models is realised, but the intermediate ``token" stage serves to normalize signal amplitude, largely removing the dependence on stimulus contrast.

Moeckel \cite{52Moeckel} presented a linear array relying on the FS token method with improved robustness to noise allowing the chip to extract motion over 2 decades of speeds.

Bartolozzi \cite{36Bartolozzi} recently presented a prototype linear array motion tracking chip which relies on the FS token method using temporal contrast edges as the token. Temporal derivatives are computed as part of the token generation process, but the chip also computes spatial derivatives in parallel. Both these derivatives as well as the motion estimates themselves are fed into a WTA circuit with programmable input weights, allowing the user to track the most salient feature in the array. The programmability of the WTA allows the most salient feature to be defined as a weighted summation the spatial contrast, temporal contrast, and motion features.

 oubieu \cite{99roubieu2013two} presented a 23.3mm $\times$ 12.3mm sensor weighing under 1 gram (including optics). The sensor consists of 5 pairs of 1D motion sensors which use tokens to measure the time for a feature to travel between neighbouring locations, similar to the Trigger and Inhibit (TI) token method.

The authors \cite{orchard2013spikingMotion} proposed an algorithm in which simple spiking neurons with pre-programmed synaptic delays can be combined with a silicon retina \cite{ATIS} to implement motion sensitive receptive fields. Similarly to Fig.~\ref{fig:FourierFrequency}a, where a point with motion in 1 spatial dimension traces out a line in a 2D space-time plot, an edge with motion in 2 spatial dimensions will trace out plane in a 3D space-time plot, with the slope of the plane encoding the local motion velocity. This can be seen as a Reichardt detector, where the inter-pixel delays uniquely describe a plane, although the algorithm still computes on temporal contrast tokens provided by the front end sensor. This approach is elaborated on in Section~\ref{sec:SAVME}.

In the author's implementation each neuron is designed to detect the presence of a particular local space-time plane (i.e. a specific inter-pixel delay) but combines responses from 5x5 pixels per motion unit to simultaneously determine both the direction and speed of the normal flow. The responses of multiple motion units are then combined in a second layer of the neural network to attenuate errors due to the aperture problem. However, this algorithm has not yet been implemented on embedded hardware or in real-time.

At a similar time, Benosman \cite{RyadFlow} proposed an approach which also relies on detection of local planes in data provided by a silicon retina. Instead of using multiple receptive fields tuned to detect different motions (planes), the best fit for a single local plane is mathematically computed, with the normal of the computed fit indicating the normal flow locally. The algorithm runs in JAVA on a host computer.

\subsection{Frequency Methods}

Others have focused on Adelson-Bergen type models \cite{Adelson:85} relying on spatiotemporal filtering for motion detection. The first such implementation was reported by Etienne-Cummings \cite{35EtienneCummings}, using a front-end silicon retina to compute a binary map of spatial edges, thereby providing an input signal of normalized amplitude. Subsequent processing using a multi-chip reconfigurable neural processor 
implements pairs of spatial and temporal filters to extract the oriented energy at a particular spatiotemporal frequency, thereby implementing an Adelson-Bergen motion unit. The oriented energy from multiple motion units of different frequencies are computed in parallel and their outputs are combined (as described in \cite{35EtienneCummings}) to obtain an estimate of image motion.

Ozalevi \cite{30Ozalevli}, in an approach similar to his implementation of the Hassenstein-Reichardt and Barlow-Levick models, described a multi-chip Adelson-Bergen model. Although successful, this model only implemented a single motion energy unit per pixel, therefore indicating the presence of a preferred motion stimulus and direction without indicating speed.

Modern computing technologies allows for processing on a larger scale than ever before. The author \cite{orchard2013SpatioTemporalMotionFPGA} demonstrated how an FPGA can be used to implement and combine 720 Adelson-Bergen motion energy units per pixel in real-time for a 128x128 pixel array running at 30FPS.

\section{A Spiking Neural Network for Visual Motion Estimation}
\label{sec:SAVME}
As mentioned in the previous section, the authors have developed a spiking neural network architecture for visual motion estimation \cite{orchard2013spikingMotion} which relies on synaptic delays to create motion sensitive receptive fields.

Discrete temporal contrast tokens from a separate front end sensor \cite{ATIS} are used as input spikes to the architecture and Leaky Integrate and Fire (LIF) neurons with a linear decay are used for computation. Such neurons are good at detecting temporal coincidence of their inputs, but motion signals are inherently spread over time, as modelled in \eqref{eq:plane}.

As stated in \cite{orchard2013spikingMotion} and repeated here for convenience, a motion sensitive unit in the architecture relies on the assumption that if we consider a small enough spatial region, a moving edge can be approximated as being a straight edge moving with constant velocity. The equation below shows how a motion stimulus is modelled.

\begin{figure}
\centering
  \includegraphics[width=0.45\textwidth]{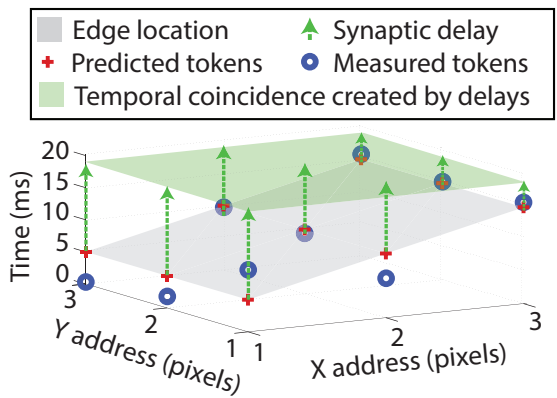}
    \caption{ Construction of a 3$\times$3 pixel receptive field sensitive to motion of an edge parallel to the y-axis travelling in the positive x-direction with speed $v_x = 1/5$ pixels per millisecond. The location of the edge can be described by $x = v_xt$, where $x$ is measured in pixels and $t$ is measured in milliseconds. This equation describes a spatiotemporal plane (shown in gray). Red crosses are located where the plane crosses pixel locations and indicate which pixels are expected to respond when (blue circles indicate actual recorded data). The length of green arrows above each pixel location indicate the synaptic delay for the synapse connecting from that pixel, and the green plane indicates the time at which the neuron would respond to this stimulus.}\label{fig:stimulus}
\end{figure}

\begin{equation}
\begin{array}{l l}
I(x,y,t) &= H(x-v_xt)\\
\frac{dI(x,y,t)}{dt} &= \delta(t-\frac{x}{v_x})\\
E(x,y,t) &= \delta(t-\frac{x}{v_x})III_1(x) III_1(y)\\
\end{array}\label{eq:plane}
\end{equation}
where $x$ and $y$ describe a location on the image plane. $I$ is intensity, $t$ is time (milliseconds), $H$ is the Heaviside step function, $v_x$ is the x-component of velocity in pixels per millisecond, $\delta$ is the Dirac delta function, $dI(x,y,t)/dt$ is the temporal derivative of image intensity, $E(x,y,t)$ is the sensor output, and $III_1$ is a sampling comb with period 1 pixel. Multiplying by the sampling combs converts the continuous space signal into a discrete space signal which only has values at integer pixel locations.

\begin{figure}
\centering
  \includegraphics[width=0.45\textwidth]{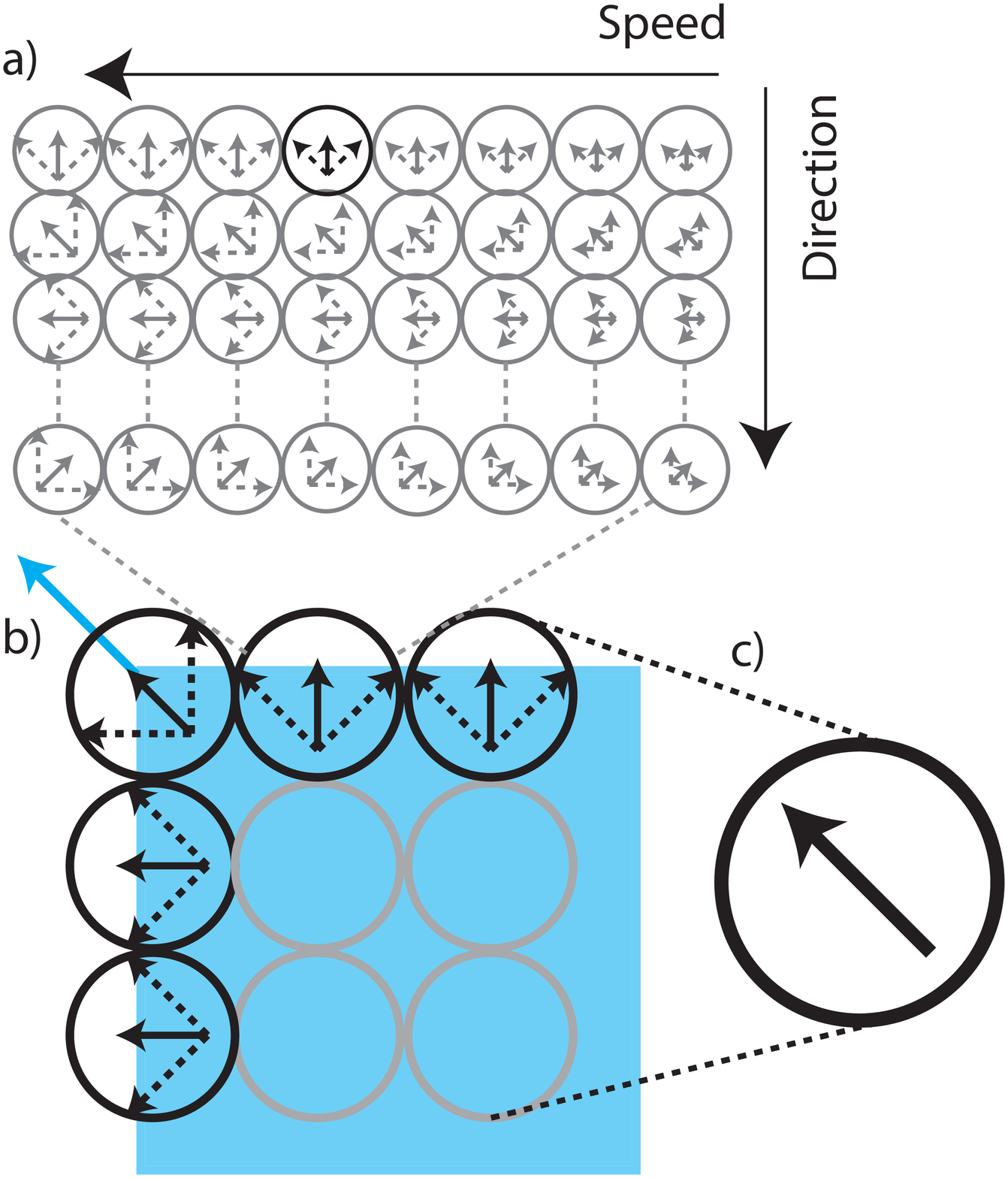}
    \caption{The multilayer architecture detecting the motion of a blue box. (a) shows a set of neurons tuned to detect different speeds and directions of motion. A full set of such neurons is present at every image location. (b) shows multiple image locations as they detect the motion of a stimulus (blue box moving upwards to the left). Black circles show locations of activated neurons, with arrows indicating which neuron (speed and direction) was activated. (c) shows a layer 2 neuron which determines the correct motion by combining layer 1 outputs to alleviate the aperture problem.}\label{fig:architecture}
\end{figure}

Fig.~\ref{fig:stimulus} shows how a receptive field sensitive to a specific motion stimulus (in this case a speed of 1/5 pixels per millisecond in the x-direction) can be constructed. The underlying concept relies on using synaptic delays (green arrows) to convert a temporal sequence of spikes (red crosses) into a group of spikes coincident in time (green plane). The delayed spikes serve as input to a LIF neuron, which is good at detecting temporal coincidence of its inputs. In practice there will not be perfect temporal coincidence because the actual spikes received from the front end sensor (blue circles) will not perfectly match the spike times predicted by our model (red crosses).

Lowering the threshold voltage of the LIF neuron will cause it to still respond when its inputs are slightly spread in time, and the threshold value can be used to control how much time spreading can be tolerated before the neuron stops responding.

Our approach can be seen as a Reichardt detector covering multiple pixels, since it is effectively delaying the signal from neighbouring pixels while the LIF neuron detects multi-pixel correlations in the delayed spikes.

As with the Reichardt detector, multiple detectors (in our case neurons) are required in order to detect different speeds and directions of motion. Our architecture uses 8$\times$8 neurons per pixel location to detect all possible combinations of 8 different directions and 8 different speeds as shown in Fig.~\ref{fig:architecture}a. Directions vary from 0 to 315 degrees in steps of 45 degrees, while speeds vary from $\sqrt{2}/{50}$ to $\sqrt{2}^8/{50}$ pixels/ms by factors of $\sqrt{2}$.

The equations in \eqref{eq:plane} are independent of motion parallel to the edge direction, presenting a form of the aperture problem where only motion perpendicular to an edge (the normal flow) can be detected. An edge moving in a direction perpendicular to its orientation at speed $s$ would look identical to the same edge moving in a direction 45 degrees to its orientation with speed $s\sqrt{2}$ (see dotted arrows in Figs.~\ref{fig:ApertureProblem}~and~\ref{fig:architecture}) since in both cases the perpendicular component of motion is just $s$. This relationship gives rise to the $\sqrt{2}$ factor used between different speeds.

A key feature which sets this work apart from previous token and Reichardt works is the use of a second stage of processing to overcome the aperture problem. The second stage of processing is implemented by a second layer of neurons, with each neuron receiving inputs over a wider spatial region than first layer neurons (Fig~\ref{fig:architecture}c). A layer 2 neuron sensitive to speed $s$ and direction $d$ would incorporate inputs from layer 1 neurons sensitive to the same speed and direction, but also from layer 1 neurons sensitive to speed $s/\sqrt{2}$ and directions $d\pm45^o$. This multi-layer approach bears resemblance to gradient based methods such as Lucas-Kanade \cite{Lucas_Kanade_1981} and Horn-Schunck \cite{Horn81determiningoptical} which compute normal flow locally in a first step before incorporating the normal flow from other nearby locations to more accurately approximate the true optical flow.

\begin{figure}
\centering
  \includegraphics[width=0.5\textwidth]{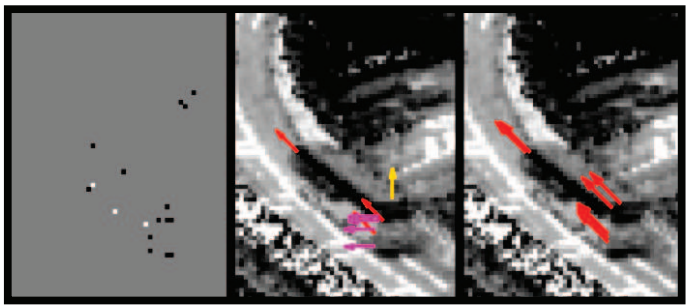}
    \caption{Motion of a moving bus as detected by the network. The left pane shows temporal change tokens (black for decrease, white for increase, grey for no change). The middle pane shows outputs of Layer 1, which tend to be perpendicular to edges. The right pane shows the output of the Layer 2, which uses the data from Layer 1 to detect the actual flow. In each pane only 3.3ms worth of data is shown. Grayscale values are obtained using the exposure measurement function of the Asynchronous Time-based Image Sensor (ATIS) \cite{ATIS}.}\label{fig:bus}
\end{figure}

Fig~\ref{fig:architecture} shows the system architecture. At each pixel location there are 8$\times8$ neurons sensitive to different speeds and directions of motion (Fig~\ref{fig:architecture}a), but mutual inhibition ensures only one neuron (shown in black) can respond to a stimulus (produce an output spike) at any time.

Fig~\ref{fig:architecture}b shows a stimulus covering 3$\times$3 pixel locations. Dark circles indicate locations where neurons are responding, with solid arrows indicating the speed and direction selectivity of the neuron responding at that location. Dotted arrows indicate other speeds and directions consistent with the aperture problem as discussed above. Fig~\ref{fig:architecture}c shows a layer 2 neuron determining the correct motion from the inputs it receives from the layer 1 neurons of Fig~\ref{fig:architecture}b.

Fig.~\ref{fig:bus} shows actual outputs from each layer for a real world scene of a bus crossing a bridge. Layer 1 outputs tending to be perpendicular to edges, while layer 2 outputs more accurately describe the actual motion of the bus by incorporating data over a larger spatial region.

\begin{figure}
\centering
  \includegraphics[width=0.5\textwidth]{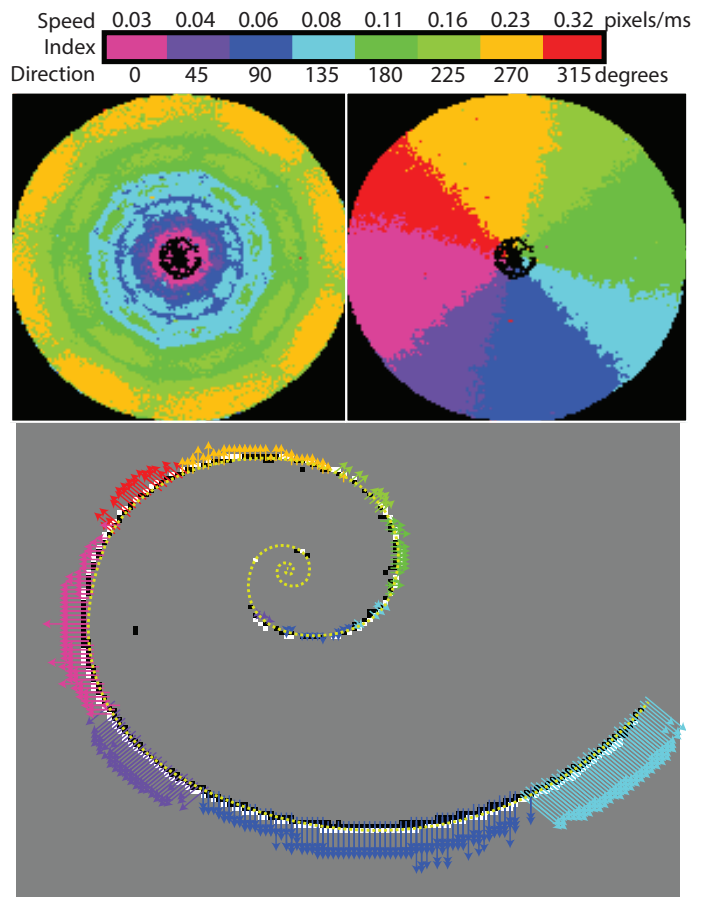}
    \caption{Output of the architecture when viewing a rotating spiral. Top images show speed (left) and direction (right) outputs accumulated over multiple rotations. The lower image shows 10ms of output data while the spiral is spinning. The motion vectors all point outwards from the center, creating a looming field.}\label{fig:spiral}
\end{figure}

Fig.~\ref{fig:spiral} presents the output of the architecture for a controlled stimulus consisting of a spinning spiral. The figure shows how it can reliably detect different speeds and directions of motion. The top images show data accumulated over multiple rotations of the spiral. Colour is used to encode speed (left) and direction (right) according to the legend provided above the images. Speed varies as a function of distance to the axis of rotation, while direction varies with angle. Near the axis of rotation, the motion is slower than the slowest receptive field and therefore elicits no responses. The lower image shows the motion architecture's output accumulated over a period of 10ms. The colour of arrows helps to encode the motion direction, while their length encodes speed. The spiral stimulus used has been superimposed on the image as a yellow dashed line.

\begin{figure}
\centering
  \includegraphics[width=0.5\textwidth]{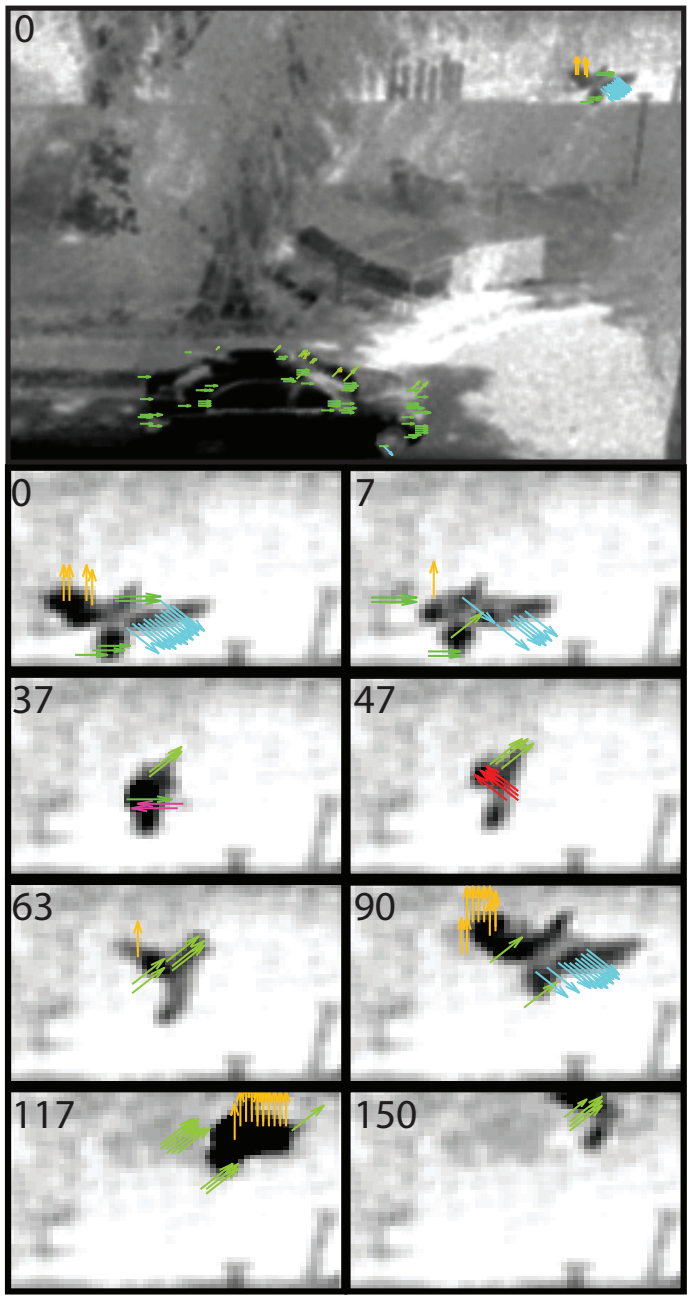}
    \caption{An example real world scene (top) consisting of a moving car and a startled bird. The lower subimages show a cropped region around the bird during flight at different points in time. The time in milliseconds is indicated at the top left of each image. Notably, the motion of each wing is detected as well as the motion of the body.}\label{fig:bird}
\end{figure}

Fig.~\ref{fig:bird} shows another example of the architecture operating on real world data. The top part of the image shows a full scene captured with the ATIS. There are two moving objects in the scene, a rightward moving car in the lower part, and a bird in the top right. The rest of the images in Fig.~\ref{fig:bird} show motion responses elicited by the flying bird. Inset numbers indicate the time (in milliseconds) at which each image was captured. Each image shows 3.3ms of motion output data. In the first frame (t~=~0) rightward motion of the bird's body is detected, while separate motion is detected for each wing. In the second frame (t~=~7ms) retraction of the left wing is detected, followed by retraction of the right wing at 37ms, as shown by the opposite motion estimates for each wing while the body continues motion upward to the right. 10ms later extension of the left wing is detected (red arrows at t~=~47ms). By 90ms, the bird has returned to a similar pose to that seen at 0ms. At 117ms, both wings have been pulled in front of the body, causing motion upwards. After 150ms, the bird exits the scene.

The bird example is particularly tricky for conventional motion estimation techniques due to the bird significantly and rapidly changing its appearance while moving.

The model as presented here has not yet been integrated into a real-time implementation. However, the model is computationally efficient. Neurons are only updated when an input spike arrives, and each neuron update only requires 4 additions, 2 \textit{greater-than} comparisons, and 1 multiplication. Sparsity of the incoming spikes from the ATIS front-end sensor keeps the required number of operations per second low and easily achievable with commercially available hardware.

The challenge in achieving real-time implementation does not lie in the computation rate, but rather in the implementation of synaptic delays. Incoming spikes must be delayed and stored, which drives up memory requirements. These memory requirements can be reduced by observing that only a small percentage of synapses are active at any one time, so memory can be shared between synapses. However, different synapses have different delays, so a simple First In First Out (FIFO) buffer will not work because the order in which spikes are placed in the buffer will not be the same as the order in which they must be read out.

\section{Future Directions}
In the previous sections we have summarized past and present works. In this section we outline the directions for future works.

Although great progress has been made, the organisation of computational circuits in artificial approaches still differs significantly from biology. In the retina, neurons are arranged in interconnected layers stacked on top of each other and lying above the photoreceptor layer. Similarly, in visual cortex neurons are arranged in 3D layers, with both short local connections, and longer range axonal connections between more distant regions of cortex.

In silicon, photoreceptors and computational circuits mimicking different layers of biological processing are restricted to lying side by side within the same plane, which limits both the photoreceptor size and spacing, which in turn affects the signal strength and spatial resolution respectively. When mapping a 3D biological structure onto 2D silicon, short vertical connections are often mapped to long lateral connections, increasing line capacitance, energy consumption, and occupying valuable space. This is overcome in many artificial implementations by only considering motion in one lateral direction, then stacking circuits in the other lateral direction instead of vertically \cite{1Tanner, 3Andreou, 4Horiuchi, 11Kramer, 19Yakovleff, 24Liu, 27Harrison, 43Sarpeshkar, 53Moeckel, 36Bartolozzi}.

As 3D stacked silicon technology matures, it can be used to alleviate the wiring problem, allow for larger photodiode fill-factors, and achieve a more biologically realistic organisation of neural circuits. This compact 3D organisation is typical of insect vision and primate retina, however, the early stages of primate retina and visual cortex are located far from each other, and thus compact integration of cortical circuits and photoreceptors is not necessarily accurate to biology.

Some of the described works have relied on an approach in which a spiking ``silicon retina" and neural processing are implemented in separate chips \cite{DVStobi, ATIS} (although 3D integration is useful for both chips). Implementing retinal and cortical processing as two different components provides advantages during system development. First, an improvement in either component can be achieved without re-fabricating the other, and second, data can be recorded as it is transmitted between the two, allowing for in depth off-line analysis which can provide insights into how neural algorithms for processing can be further improved.

The last decade has seen silicon retinae mature to the point where they are now commercially available and are used by many labs around the world and there are a number of works emerging (see other papers in this special issue) which argue for the superiority of these sensors for high speed visual tasks which must be executed in real-time on a limited power budget. Reconfigurable neural processing platforms are also rapidly maturing, spurred by the dramatic increase in interest and funding the neuromorphic field has experienced in the last few years.

It is an exciting time for the neuromorphic area, with major companies including Qualcomm, Samsung, Intel, and IBM coming on board and launching their own research projects in the field. There are also major projects in the US (the \$200 million BRAIN initiative \cite{BRAIN}) and Europe (the \texteuro1 billion Human Brain Project \cite{HBP}) which are incorporating neuromorphic aspects. The US Defense Advanced Research Projects Agency (DARPA) has also been taking notice, funding projects such as the \textit{Unconventional Processing Of Signals For Intelligent Data Exploitation} (UPSIDE) and \textit{Systems of Neuromorphic Adaptive Plastic Scalable Electronics} (SYNAPSE) projects.

Modern CMOS technology is quite different from biological ``wetware". CMOS typically operates at frequencies ranging from megahertz to gigahertz, while a general rule of thumb is that biological neurons do not operate at frequencies above 1kHz. This massive speed difference is not necessarily an advantage for silicon. In fact, slowing silicon neural circuits down to biologically realistic time-scales can prove quite challenging, and often requires extra design effort and cost to implement.

Biology leverages parallel processing, and the speed of CMOS can be useful when one wants to approximate multiple parallel units from biology using a single high speed sequential unit in silicon. However, this approach comes at a disproportionate power cost. Higher operating speeds require higher operating voltages, and power scales proportionally to voltage squared. It is therefore preferable to have many low speed, low voltage processors (like biology) than a few high speed, high voltage processors (like modern CMOS). Hence, biology provides a road map for the future, where the scaling of CMOS will allow the realization of ultra-low voltage (hence low-power) circuits performing massively parallel computation in very small and three dimensionally stacked dies. As technology moves in this direction, CMOS can learn about 3D connectivity, massively parallel computation, density of computational elements, and stochastic circuits from biology.

Despite technological improvements, wiring remains an issue in the connectivity which can currently be achieved between artificial neurons. Even though 3D integration can help, inter-neuron connectivity with present technologies is constrained to remain far sparser than in biology. The combined use of silicon circuits and carbon nanotube crossbar arrays has been proposed to improve physical connectivity, with memristor devices capable of learning proposed for use as synaptic connections between nanotubes \cite{serrano2013proposal}.

The development of new online learning algorithms and architectures, whether relying on memristor devices or conventional silicon, are likely to play an increasingly important role. Using learning through visual experience to help configure and organise a neural architecture can improve fault tolerance (and therefore device yield) and save man hours spent on manual configuration. This learning is especially important if copies of the same device are to adapt to operation under very different visual conditions, such as in urban versus forested environments, or onboard flying versus ground vehicles. 

As mentioned in the Introduction, biological sensors are embodied, and have evolved in conjunction with motor systems. The interplay between motor and sensory systems can be useful for sensing. Examples of this include the peering behaviour used by many animals to induce motion parallax for depth perception \cite{MotionParallax}, the optomotor response in insects \cite{reichardt1956geschwindigkeitsverteilung}, and the vestibular ocular reflex in humans \cite{VestibularOcularReflex}. Recent studies also suggest that micro-saccades during fixation play an important role in perception, particularly for object recognition in humans \cite{microsaccades}. In \textit{Drosophila} motion is found to also play a role in motion perception \cite{Franceschini}.

An embodied biological sensor also serves a particular purpose, to provide information relevant to the agent for self-preservation and meaningful interaction with the environment. The value metric of biological motion estimates is therefore not directly assessed by how accurately motion is perceived, but rather by how motion estimates improve the effectiveness of the agent's behaviour (although to a degree, more accurate motion estimates will be more effective in affecting behaviour).

It is therefore important to keep in mind the intended use of the system being constructed. Inspiration from biology is useful, but at some stage the design must deviate from precise bio-mimicry. A micro-aerial vehicle may benefit from an artificial version of the \textit{Drosophila} vision system, but for the vehicle to be of value to the operator, it will be expected to execute a goal oriented task rather than simply behave like \textit{Drosophila}. Also, at some point making a system more biologically accurate will come at a performance cost rather than benefit due to the inherent differences between silicon circuits and biological neurons. It was Carver Mead who first developed the concept of imitating neural processing in silicon circuits by noting the similarities between the two \cite{NeuromorphicSytems}, but it was also Carver Mead who said ``Listen to the technology; find out what it's telling you".

Nevertheless, we are still a long way from matching the power efficient performance of biology in artificial systems, so for the foreseeable future, continued research into bio-inspired visual motion estimation techniques will reap rewards for artificial systems.

\bibliographystyle{unsrt}
\bibliography{0069-SIP-2014-PIEEE}

\begin{IEEEbiography}[{\includegraphics[width=1in,height=1.25in,clip,keepaspectratio]{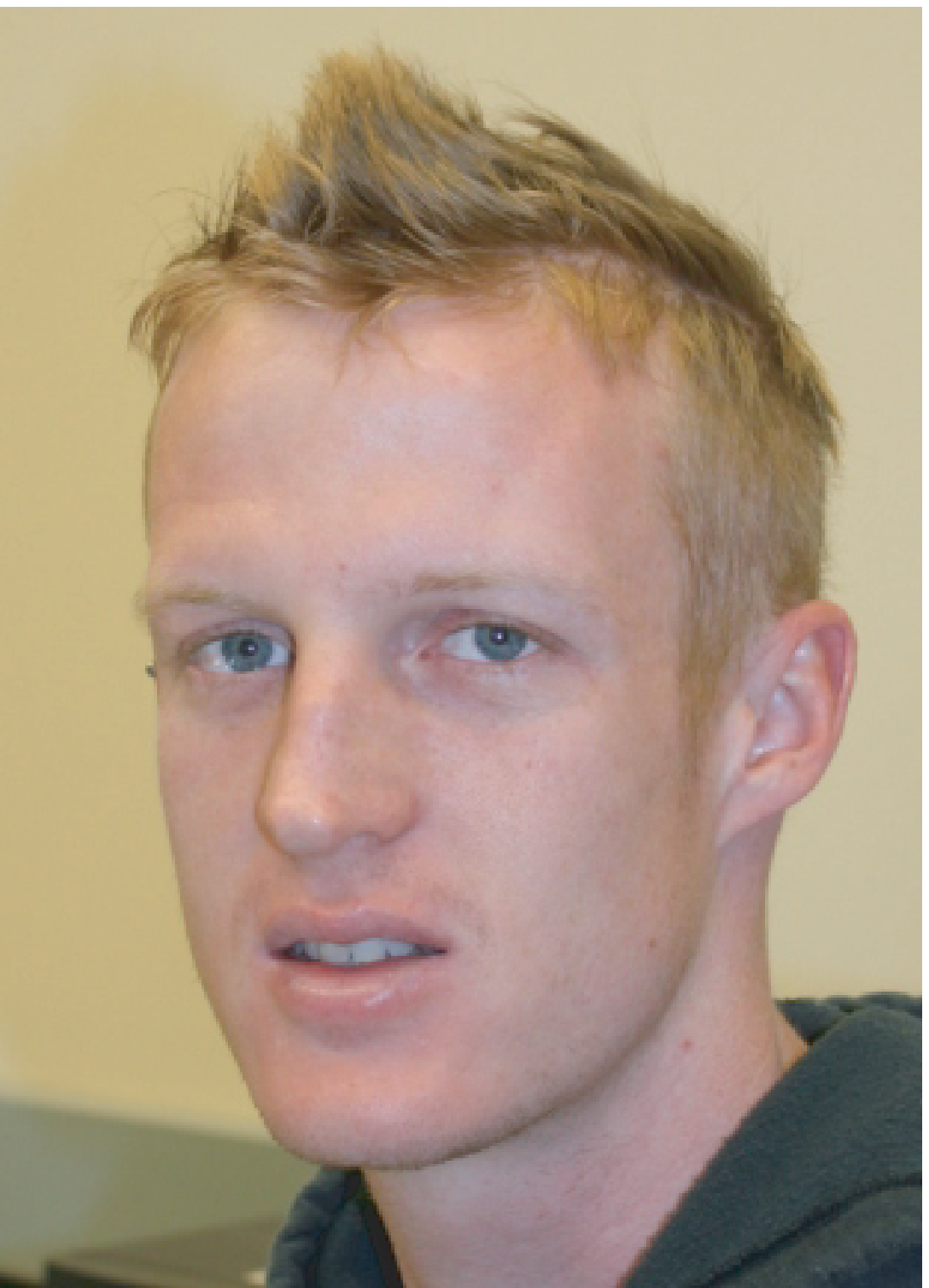}}]{Garrick Orchard}\input{Biographies/Orcha.txt}
\end{IEEEbiography}

\begin{IEEEbiography}[{\includegraphics[width=1in,height=1.25in,clip,keepaspectratio]{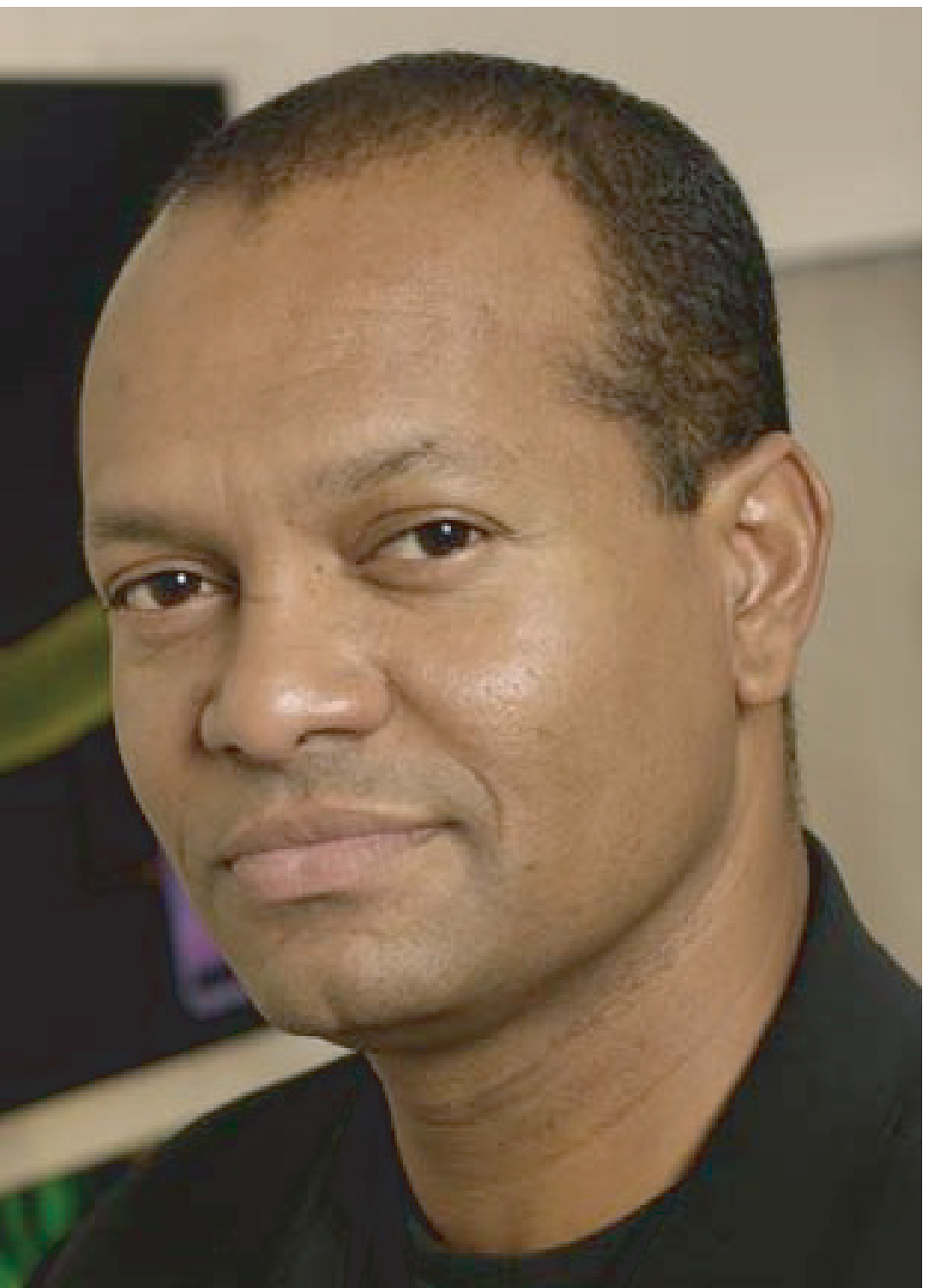}}]{Ralph Etienne-Cummings}\input{Biographies/Etien.txt}
\end{IEEEbiography}

\end{document}

%% file: Biographies/Orcha.txt
received the B.Sc. degree in electrical engineering from the University of Cape Town, South Africa in 2006 and the M.S.E. and Ph.D. degrees in electrical and computer engineering from Johns Hopkins University, Baltimore in 2009 and 2012 respectively. He was named a Paul V. Renoff fellow in 2007 and a Virginia and Edward M. Wysocki, Sr. fellow in 2011, was a recipient of the JHUAPL Hart Prize for Best Research and Development Project, and won the best live demonstration prize at the IEEE Biocas 2012 conference. He is currently a Postdoctoral Research Fellow at the Singapore Institute for Neurotechnology (SINAPSE) at the National University of Singapore where his research focuses on developing neuromorphic vision algorithms and systems for real-time sensing on aerial platforms. His other research interests include mixed-signal very large scale integration (VLSI) design, compressive sensing, spiking neural networks, visual motion perception, and legged locomotion.

%% file: Biographies/Etien.txt
received his B. Sc. in physics, 1988, from Lincoln University, Pennsylvania. He completed his M.S.E.E. and Ph.D. in electrical engineering at the University of Pennsylvania in 1991 and 1994, respectively. He is currently a professor of electrical and computer engineering, and computer science at Johns Hopkins University (JHU). He is the former Director of Computer Engineering at JHU and the Institute of Neuromorphic Engineering. He is also the Associate Director for Education and Outreach of the National Science Foundation (NSF) sponsored Engineering Research Centers on Computer Integrated Surgical Systems and Technology at JHU. He has served as Chairman of the IEEE Circuits and Systems (CAS) Technical Committee on Sensory Systems and on Neural Systems and Application. He was also the General Chair of the IEEE BioCAS 2008 Conference. He was also a member of Imagers, MEMS, Medical and Displays Technical Committee of the ISSCC Conference from 1999 – 2006. He is the recipient of the NSF's Career and Office of Naval Research Young Investigator Program Awards. In 2006, he was named a Visiting African Fellow and a Fulbright Fellowship Grantee for his sabbatical at University of Cape Town, South Africa. He was invited to be a lecturer at the National Academies of Science Kavli Frontiers Program, in 2007. He has won publication awards including the 2003 Best Paper Award of the EURASIP Journal of Applied Signal Processing and "Best Ph.D. in a Nutshell" at the IEEE BioCAS 2008 Conference, and has been recognized for his activities in promoting the participation of women and minorities in science, technology, engineering and mathematics. His research interest includes mixed signal VLSI systems, computational sensors, computer vision, neuromorphic engineering, smart structures, mobile robotics, legged locomotion and neuroprosthetic devices.

%% file: 0069-SIP-2014-PIEEE.bbl
\begin{thebibliography}{100}

\bibitem{lee1980optic}
D.~N. Lee and H.~Kalmus.
\newblock The optic flow field: The foundation of vision.
\newblock {\em Philosophical Trans. of the Royal Society of London. B,
  Biological Sciences}, 290(1038):169--179, 1980.

\bibitem{srinivasan2004visual}
M.~V. Srinivasan and S.~Zhang.
\newblock Visual motor computations in insects.
\newblock {\em Annu. Rev. Neurosci.}, 27:679--696, 2004.

\bibitem{1Tanner}
J.~E. Tanner.
\newblock {\em Integrated optical motion detection}.
\newblock PhD thesis, California Institute of Technology, CA, USA, 1986.

\bibitem{WolpertTED}
D.~Wolpert.
\newblock The real reason for brains.
\newblock {\em TED Talks}, July 2011.

\bibitem{StatisticsOfNaturalImages}
A.~Hyvrinen, J.~Hurri, and P.~O. Hoyer.
\newblock {\em Natural Image Statistics: A Probabilistic Approach to Early
  Computational Vision}.
\newblock Springer Publishing Company, Incorporated, 1st edition, 2009.

\bibitem{MousePatentAdan}
M.~E. Adan, T.~Aoyagi, T.~E. Holmdahl, T.~M. Lipscomb, and T.~Miura.
\newblock Operator input device, Jan 2001.
\newblock US Patent 6,172,354.

\bibitem{MousePatentHartlove}
G.~B. Gordon, D.~L. Knee, R.~Badyal, and J.~T. Hartlove.
\newblock Seeing eye mouse for a computer system, Aug 2002.
\newblock US Patent 6,433,780.

\bibitem{MousePatentTanner}
C.~A. Mead and J.~E. Tanner.
\newblock Correlating optical motion detector, Dec 1986.
\newblock US Patent 4,631,400.

\bibitem{px4flowBlockMatching}
D.~Honegger, L.~Meier, P.~Tanskanen, and M.~Pollefeys.
\newblock An open source and open hardware embedded metric optical flow {CMOS}
  camera for indoor and outdoor applications.
\newblock In {\em Robotics and Automation, IEEE Int. Conf. on (ICRA)}, pages
  1736--1741, 2013.

\bibitem{SparseCoding}
B.~A. Olshausen and D.~J. Field.
\newblock Sparse coding of sensory inputs.
\newblock {\em Current opinion in neurobiology}, 14(4):481--487, 2004.

\bibitem{BOLDsignal}
N.~K. Logothetis and B.~A. Wandell.
\newblock Interpreting the {BOLD} signal.
\newblock {\em Annu. Rev. Physiol.}, 66:735--769, 2004.

\bibitem{DVStobi}
P.~Lichtsteiner, C.~Posch, and T.~Delbruck.
\newblock A 128x128 120 {dB} 15 us latency asynchronous temporal contrast
  vision sensor.
\newblock {\em Solid-State Circuits, IEEE J. of}, 43(2):566--576, Feb 2008.

\bibitem{ATIS}
C.~Posch, D.~Matolin, and R.~Wohlgenannt.
\newblock A {QVGA} 143 {dB} dynamic range frame-free {PWM} image sensor with
  lossless pixel-level video compression and time-domain {CDS}.
\newblock {\em Solid-State Circuits, IEEE J. of}, 46(1):259--275, Jan 2011.

\bibitem{NeuroGrid}
B.~V. Benjamin, P.~Gao, E.~McQuinn, S.~Choudhary, A.~R. Chandrasekaran, J.-M.
  Bussat, R.~Alvarez-Icaza, J.~V. Arthur, P.~A Merolla, and K.~Boahen.
\newblock Neurogrid: {A} mixed-analog-digital multichip system for large-scale
  neural simulations.
\newblock {\em Proc. of the IEEE}, 102(5):699--716, May 2014.

\bibitem{TemporalPrecision}
S.~Panzeri, N.~Brunel, N.~K. Logothetis, and C.~Kayser.
\newblock Sensory neural codes using multiplexed temporal scales.
\newblock {\em Trends in Neurosciences}, 33(3):111--120, 2010.

\bibitem{Spinnaker}
E.~Painkras, L.~A. Plana, J.~Garside, S.~Temple, F.~Galluppi, C.~Patterson,
  D.~R. Lester, A.~D. Brown, and S.~B. Furber.
\newblock {SpiNNaker}: {A} 1-{W} 18-core system-on-chip for massively-parallel
  neural network simulation.
\newblock {\em Solid-State Circuits, IEEE J. of}, 2013.

\bibitem{NanotechnologyYield}
G.~A. Silva.
\newblock Neuroscience nanotechnology: progress, opportunities and challenges.
\newblock {\em Nature Reviews Neuroscience}, 7(1):65--74, 2006.

\bibitem{OcularDominanceSuture}
T.~N. Wiesel and D.~H. Hubel.
\newblock Single-cell responses in striate cortex of kittens deprived of vision
  in one eye.
\newblock {\em J. Neurophysiol}, 26(6):1003--1017, 1963.

\bibitem{EnvironmentEffect}
T.~N. Wiesel.
\newblock The postnatal development of the visual cortex and the influence of
  environment.
\newblock {\em Bioscience reports}, 2(6):351--377, 1982.

\bibitem{jahne2000computer}
B.~Jahne.
\newblock {\em Computer vision and applications: a guide for students and
  practitioners}.
\newblock Academic Press, 2000.

\bibitem{aperture1988nakayama}
K.~Nakayama and G.~H. Silverman.
\newblock The aperture problem. perception of nonrigidity and motion direction
  in translating sinusoidal lines.
\newblock {\em Vision research}, 28(6):739--746, 1988.

\bibitem{Lucas_Kanade_1981}
B.~D. Lucas and T.~Kanade.
\newblock An iterative image registration technique with an application to
  stereo vision.
\newblock {\em Proc. DARPA Image Understanding Workshop}, pages 121--130, Apr
  1981.

\bibitem{Horn81determiningoptical}
B.~K.~P. Horn and B.~G. Schunck.
\newblock Determining optical flow.
\newblock {\em Artificial Intelligence}, 17:185--203, 1981.

\bibitem{Adelson:85}
E.~H. Adelson and J.~R. Bergen.
\newblock Spatiotemporal energy models for the perception of motion.
\newblock {\em J. Opt. Soc. Am.}, 2(2):284--299, 1985.

\bibitem{dirac1981principles}
P.~A.~M. Dirac.
\newblock {\em The principles of quantum mechanics}.
\newblock Oxford University Press, 1981.

\bibitem{Fourier}
R.~N. Bracewell.
\newblock {\em The Fourier transform and its applications}.
\newblock {McGraw-Hill Education, New York}, 1980.

\bibitem{39AlanStockerMain}
A.~Stocker.
\newblock Analog integrated 2-{D} optical flow sensor.
\newblock {\em Analog Integrated Circuits and Signal Processing}, 46:121--138,
  2006.

\bibitem{32Higgins}
C.~M. Higgins, V.~Pant, and R.~Deutschmann.
\newblock Analog {VLSI} implementation of spatio-temporal frequency tuned
  visual motion algorithms.
\newblock {\em Circuits and Systems I: Regular Papers, IEEE Trans. on},
  52(3):489--502, March 2005.

\bibitem{EyeEvolution}
D.-E. Nilsson.
\newblock Eye ancestry: old genes for new eyes.
\newblock {\em Current Biology}, 6(1):39--42, 1996.

\bibitem{AnimalEyes}
M.~F. Land and D.-E. Nilsson.
\newblock {\em Animal eyes}.
\newblock Oxford University Press, 2012.

\bibitem{borst2010fly}
A.~Borst, J.~Haag, and D.~F. Reiff.
\newblock Fly motion vision.
\newblock {\em Annual review of neuroscience}, 33:49--70, 2010.

\bibitem{martin2011information}
P.~R. Martin and S.~G. Solomon.
\newblock Information processing in the primate visual system.
\newblock {\em The J. of physiology}, 589(1):29--31, 2011.

\bibitem{FunctionalSpecialization}
M.~Joesch, F.~Weber, H.~Eichner, and A.~Borst.
\newblock Functional specialization of parallel motion detection circuits in
  the fly.
\newblock {\em J. of Neuroscience}, 33(3):902--905, 2013.

\bibitem{RoleOfStarburstAmacrine}
W.~R. Taylor and R.~G. Smith.
\newblock The role of starburst amacrine cells in visual signal processing.
\newblock {\em Visual neuroscience}, 29(01):73--81, 2012.

\bibitem{drosophilaVision}
M.~Joesch, B.~Schnell, S.~V. Raghu, D.~F. Reiff, and A.~Borst.
\newblock On and off pathways in {D}rosophila motion vision.
\newblock {\em Nature}, 468(7321):300--304, 2010.

\bibitem{reichardt1961autocorrelation}
W.~Reichardt.
\newblock Autocorrelation, a principle for the evaluation of sensory
  information by the central nervous system.
\newblock {\em Sensory communication}, pages 303--317, 1961.

\bibitem{CortexMotion}
J.~H. Maunsell and D.~C. Van~Essen.
\newblock Functional properties of neurons in middle temporal visual area of
  the macaque monkey. {I}. selectivity for stimulus direction, speed, and
  orientation.
\newblock {\em J Neurophysiol}, 49(5):1127--1147, 1983.

\bibitem{MotionMT}
W.~T. Newsome and E.~B. Pare.
\newblock A selective impairment of motion perception following lesions of the
  middle temporal visual area ({MT}).
\newblock {\em The J. of Neuroscience}, 8(6):2201--2211, 1988.

\bibitem{SeeingThings}
A.~Borst and T.~Euler.
\newblock Seeing things in motion: models, circuits, and mechanisms.
\newblock {\em Neuron}, 71(6):974--994, 2011.

\bibitem{VestibularOcularReflex}
D.~E. Angelaki and K.~E. Cullen.
\newblock Vestibular system: the many facets of a multimodal sense.
\newblock {\em Annu. Rev. Neurosci.}, 31:125--150, 2008.

\bibitem{SaccadeSuppression}
B.~G. Breitmeyer and L.~Ganz.
\newblock Implications of sustained and transient channels for theories of
  visual pattern masking, saccadic suppression, and information processing.
\newblock {\em Psychological review}, 83(1):1, 1976.

\bibitem{HumanVisualMotor}
A.~D. Milner, M.~A. Goodale, and A.~J. Vingrys.
\newblock {\em The visual brain in action}, volume~2.
\newblock Oxford University Press, 2006.

\bibitem{optomotorFly}
K.~Hausen.
\newblock Motion sensitive interneurons in the optomotor system of the fly.
\newblock {\em Biological Cybernetics}, 45(2):143--156, 1982.

\bibitem{HUBELWeisel}
D.~H. Hubel and T.~N. Weisel.
\newblock Receptive fields, binocular interaction, and function architecture in
  the cat's visual cortex.
\newblock {\em J Physiol (Lond.)}, 160:106--154, 1962.

\bibitem{barlow1965mechanism}
H.~B. Barlow and W.~R. Levick.
\newblock The mechanism of directionally selective units in rabbit's retina.
\newblock {\em The J. of physiology}, 178(3):477, 1965.

\bibitem{Watson:85}
A.~B. Watson and A.~J.~Jr. Ahumada.
\newblock Model of human visual-motion sensing.
\newblock {\em J. Opt. Soc. Am. A}, 2(2):322--341, 1985.

\bibitem{Fermuller2010315}
C.~Fermuller, H.~Ji, and A.~Kitaoka.
\newblock Illusory motion due to causal time filtering.
\newblock {\em Vision Research}, 50(3):315--329, 2010.

\bibitem{illusionsweiss2002motion}
Y.~Weiss, E.~P. Simoncelli, and E.~H. Adelson.
\newblock Motion illusions as optimal percepts.
\newblock {\em Nature neuroscience}, 5(6):598--604, 2002.

\bibitem{SimoncelliMT}
E.~P. Simoncelli and D.~J. Heeger.
\newblock Representing retinal image speed in visual cortex.
\newblock {\em Nature Neuroscience}, pages 461--462, 2001.

\bibitem{simoncelli2003local}
E.~P. Simoncelli.
\newblock Local analysis of visual motion.
\newblock {\em The visual neurosciences}, pages 1616--1623, 2003.

\bibitem{35EtienneCummings}
R.~Etienne-Cummings, J.~Van~der Spiegel, and P.~Mueller.
\newblock Hardware implementation of a visual-motion pixel using oriented
  spatiotemporal neural filters.
\newblock {\em Circuits and Systems II: Analog and Digital Signal Processing,
  IEEE Trans. on}, 46(9):1121--1136, Sep 1999.

\bibitem{Franceschini}
N.~Franceschini.
\newblock Small brains, smart machines: From fly vision to robot vision and
  back again.
\newblock {\em Proc. of the IEEE}, 102(5):751--781, May 2014.

\bibitem{3Andreou}
A.~G. Andreou, K.~Strohbehn, and R.~E. Jenkins.
\newblock Silicon retina for motion computation.
\newblock In {\em Circuits and Systems, IEEE Int. Sympoisum on (ISCAS)}, pages
  1373--1376 vol.3, Jun 1991.

\bibitem{6EtienneCummings}
R.~Etienne-Cummings, S.~A. Fernando, J.~Van~der Spiegel, and P.~Mueller.
\newblock Real-time 2{D} analog motion detector {VLSI} circuit.
\newblock In {\em Neural Networks, Int. Joint Conf. on (IJCNN)}, volume~4,
  pages 426--431, Jun 1992.

\bibitem{4Horiuchi}
T.~Horiuchi, J.~Lazzaro, A.~Moore, and C.~Koch.
\newblock A delay-line based motion detection chip.
\newblock In {\em Adv. Neural Inf. Proc. Systems}, pages 406--412, 1991.

\bibitem{7Horiuchi}
T.~Horiuchi, W.~Bair, B.~Bishofberger, A.~Moore, C.~Koch, and J.~Lazzaro.
\newblock Computing motion using analog {VLSI} vision chips: An experimental
  comparison among different approaches.
\newblock {\em Int. J. of Computer Vision}, 8(3):203--216, 1992.

\bibitem{8Delbruck}
T.~Delbruck.
\newblock Silicon retina with correlation-based, velocity-tuned pixels.
\newblock {\em Neural Networks, IEEE Trans. on}, 4(3):529--541, May 1993.

\bibitem{43Sarpeshkar}
R.~Sarpeshkar, W.~Bair, and C.~Koch.
\newblock Visual motion computation in analog {VLSI} using pulses.
\newblock In {\em Advances in Neural Information Processing Systems}, volume~5,
  pages 781--788, 1993.

\bibitem{44Sarpeshkar}
R.~Sarpeshkar, J.~Kramer, G.~Indiveri, and C.~Koch.
\newblock Analog {VLSI} architectures for motion processing: From fundamental
  limits to system applications.
\newblock In {\em Proc. IEEE}, pages 969--987, 1996.

\bibitem{40Gottardi}
M.~Gottardi and W.~Yang.
\newblock A {CCD/CMOS} image motion sensor.
\newblock In {\em Solid-State Circuits Conf., Digest of Technical Papers. IEEE
  Int. (ISSCC)}, pages 194--195, Feb 1993.

\bibitem{10Kramer}
J.~Kramer.
\newblock Compact integrated motion sensor with three-pixel interaction.
\newblock {\em Pattern Analysis and Machine Intelligence, IEEE Trans. on},
  18(4):455--460, Apr 1996.

\bibitem{11Kramer}
J.~Krammer and C.~Koch.
\newblock Pulse-based analog {VLSI} velocity sensors.
\newblock {\em Circuits and Systems II: Analog and Digital Signal Processing,
  IEEE Trans. on}, 44(2):86--101, Feb 1997.

\bibitem{19Yakovleff}
A.~J.~S. Yakovleff and A.~Moini.
\newblock Motion perception using analog {VLSI}.
\newblock {\em Analog Integrated Circuits and Signal Processing},
  15(2):183--200, 1998.

\bibitem{Andrearreguit1996cmos}
X.~Arreguit, F.~A. Van~Schaik, F.~V. Bauduin, M.~Bidiville, and E.~Raeber.
\newblock A {CMOS} motion detector system for pointing devices.
\newblock {\em Solid-State Circuits, IEEE J. of}, 31(12):1916--1921, 1996.

\bibitem{14EtienneCummings}
R.~Etinne-Cummings, J.~Van~der Spiegel, P.~Mueller, and M.-Z. Zhang.
\newblock A foveated visual tracking chip.
\newblock In {\em Solid-State Circuits Conf., Digest of Technical Papers., IEEE
  Int. (ISSCC)}, pages 38--39, Feb 1997.

\bibitem{15EtienneCummings}
R.~Etienne-Cummings, J.~Van Der~Spiegel, and P.~Mueller.
\newblock A focal plane visual motion measurement sensor.
\newblock {\em IEEE Trans. Circuits Syst. II}, pages 55--66, 1997.

\bibitem{16Moini}
A.~Moini, A.~Bouzerdoum, K.~Eshraghian, A.~Yakovleff, X.~T. Nguyen,
  A.~Blanksby, R.~Beare, D.~Abbott, and R.~E. Bogner.
\newblock An insect vision-based motion detection chip.
\newblock {\em Solid-State Circuits, IEEE J. of}, 32(2):279--284, 1997.

\bibitem{moini2000vision}
A.~Moini.
\newblock {\em Vision Chips}.
\newblock Springer, 2000.

\bibitem{18Harrison}
R.~R. Harrison and K.~Christof.
\newblock An analog {VLSI} model of the fly elementary motion detector.
\newblock In {\em Advances in Neural Information Processing Systems 10}, pages
  880--886. MIT Press, 1998.

\bibitem{23Higgins}
C.~M. Higgins, R.~A. Deutschmann, and C.~Koch.
\newblock Pulse-based 2-{D} motion sensors.
\newblock {\em Circuits and Systems II: Analog and Digital Signal Processing,
  IEEE Trans. on}, 46(6):677--687, Jun 1999.

\bibitem{20Indiveri}
G.~Indiveri, A.~M. Whatley, and J.~Kramer.
\newblock A reconfigurable neuromorphic {VLSI} multi-chip system applied to
  visual motion computation.
\newblock In {\em Proc. of the Seventh Int. Conf. on Microelectronics for
  Neural, Fuzzy and Bio-inspired Systems; Microneuro'99}, pages 37--44, Los
  Alamitos, CA, April 1999. {IEEE} Computer Society.

\bibitem{21Jiang}
H.-C. Jiang and C.-Y. Wu.
\newblock A 2-{D} velocity- and direction-selective sensor with {BJT}-based
  silicon retina and temporal zero-crossing detector.
\newblock {\em Solid-State Circuits, IEEE J. of}, 34(2):241--247, Feb 1999.

\bibitem{47Barrows}
G.~L. Barrows and C.~Neely.
\newblock Mixed-mode {VLSI} optic flow sensors for in-flight control of a micro
  air vehicle.
\newblock {\em Proc. SPIE}, 4109:52--63, 2000.

\bibitem{24Liu}
S.-C. Liu.
\newblock A neuromorphic a{VLSI} model of global motion processing in the fly.
\newblock {\em Circuits and Systems II: Analog and Digital Signal Processing,
  IEEE Trans. on}, 47(12):1458--1467, Dec 2000.

\bibitem{31Pant}
V.~Pant and C.~M. Higgins.
\newblock A biomimetic {VLSI} architecture for small target tracking.
\newblock In {\em Circuits and Systems, Proc. of the Int. Symp. on (ISCAS)},
  volume~3, pages 5--8, May 2004.

\bibitem{25Higgins}
C.~M. Higgins and C.~Koch.
\newblock A modular multi-chip neuromorphic architecture for real-time visual
  motion processing.
\newblock {\em Analog Integr. Circuit. Signal}, pages 195--211, 2000.

\bibitem{27Harrison}
R.~R. Harrison and C.~Koch.
\newblock A robust analog {VLSI} {R}eichardt motion sensor.
\newblock {\em Analog Integrated Circuits and Signal Processing},
  24(3):213--229, 2000.

\bibitem{29Higgins}
C.~M. Higgins and S.~A. Shams.
\newblock A biologically inspired modular {VLSI} system for visual measurement
  of self-motion.
\newblock {\em Sensors J., IEEE}, 2(6):508--528, Dec 2002.

\bibitem{28Yamada}
K.~Yamada and M.~Soga.
\newblock A compact integrated visual motion sensor for its applications.
\newblock {\em Intelligent Transportation Systems, IEEE Trans. on},
  4(1):35--42, March 2003.

\bibitem{30Ozalevli}
E.~Ozalevli and C.~M. Higgins.
\newblock Multi-chip implementation of a biomimetic {VLSI} vision sensor based
  on the {Adelson-Bergen} algorithm.
\newblock In {\em Proc. of the Joint Int. Conf. on Artificial Neural Networks
  and Neural Information Processing}, ICANN/ICONIP, pages 433--440, Berlin,
  Heidelberg, 2003. Springer-Verlag.

\bibitem{50Massie}
M.~Massie, C.~Baxter, J.~P. Curzan, P.~McCarley, and R.~Etienne-Cummings.
\newblock Vision chip for navigating and controlling micro unmanned aerial
  vehicles.
\newblock In {\em Circuits and Systems. Proc. of the Int. Symp. on (ISCAS)},
  volume~3, pages 786--789, May 2003.

\bibitem{33Ozalevli}
E.~Ozalevli and C.~M. Higgins.
\newblock Reconfigurable biologically inspired visual motion systems using
  modular neuromorphic {VLSI} chips.
\newblock {\em Circuits and Systems I: Regular Papers, IEEE Trans. on},
  52(1):79--92, Jan 2005.

\bibitem{34Ozalevli}
E.~Ozalevli, P.~Hasler, and C.~M. Higgins.
\newblock Winner-take-all-based visual motion sensors.
\newblock {\em Circuits and Systems II: Express Briefs, IEEE Trans. on},
  53(8):717--721, Aug 2006.

\bibitem{51Harrison}
R.~R. Harrison.
\newblock A biologically inspired analog {IC} for visual collision detection.
\newblock {\em Circuits and Systems I: Regular Papers, IEEE Trans. on},
  52(11):2308--2318, Nov 2005.

\bibitem{97shoemaker2005insectVLSI}
P.~A. Shoemaker and D.~C. O'Carroll.
\newblock Insect-based visual motion detection with contrast adaptation.
\newblock In {\em Defense and Security}, pages 292--303. International Society
  for Optics and Photonics, 2005.

\bibitem{Mehta2}
S.~Mehta and R.~Etienne-Cummings.
\newblock A simplified normal optical flow measurement {CMOS} camera.
\newblock {\em Circuits and Systems I: Regular Papers, IEEE Trans. on},
  53(6):1223--1234, June 2006.

\bibitem{52Moeckel}
R.~Moeckel and S.-C. Liu.
\newblock Motion detection circuits for a time-to-travel algorithm.
\newblock In {\em Circuits and Systems, IEEE Int. Symp. on (ISCAS)}, pages
  3079--3082, May 2007.

\bibitem{36Bartolozzi}
C.~Bartolozzi, N.~K. Mandloi, and G.~Indiveri.
\newblock Attentive motion sensor for mobile robotic applications.
\newblock In {\em Circuits and Systems, IEEE Int. Symp. on (ISCAS)}, pages
  2813--2816, May 2011.

\bibitem{99roubieu2013two}
F.~L. Roubieu, F.~Expert, G.~Sabiron, and F.~Ruffier.
\newblock Two-directional 1-g visual motion sensor inspired by the fly's eye.
\newblock {\em Sensors J., IEEE}, 13(3):1025--1035, 2013.

\bibitem{AlanBook}
A.~A. Stocker.
\newblock {\em Analog {VLSI} Circuits for the Perception of Visual Motion}.
\newblock John Wiley \& Sons, 2006.

\bibitem{orchard2013spikingMotion}
G.~Orchard, R.~Benosman, R.~Etienne-Cummings, and N.V. Thakor.
\newblock A spiking neural network architecture for visual motion estimation.
\newblock In {\em IEEE Biomedical Circuits and Systems Conf. (BioCAS)}, pages
  298--301. IEEE, 2013.

\bibitem{RyadFlow}
R.~Benosman, C.~Clercq, X.~Lagorce, S.-H. Ieng, and C.~Bartolozzi.
\newblock Event-based visual flow.
\newblock {\em Neural Networks and Learning Systems, IEEE Trans. on},
  25(2):407--417, Feb 2014.

\bibitem{orchard2013SpatioTemporalMotionFPGA}
G.~Orchard, N.~V. Thakor, and R.~Etienne-Cummings.
\newblock Real-time motion estimation using spatiotemporal filtering in {FPGA}.
\newblock In {\em IEEE Biomedical Circuits and Systems Conf. (BioCAS)}, pages
  306--309. IEEE, 2013.

\bibitem{53Moeckel}
R.~Moeckel, R.~Jaeggi, and S.-C. Liu.
\newblock Steering with an a{VLSI} motion detection chip.
\newblock In {\em Circuits and Systems, IEEE Int. Symp. on (ISCAS)}, pages
  1036--1039, May 2008.

\bibitem{BRAIN}
A.~P. Alivisatos, M.~Chun, G.~M. Church, R.~J. Greenspan, M.~L. Roukes, and
  R.~Yuste.
\newblock The brain activity map project and the challenge of functional
  connectomics.
\newblock {\em Neuron}, 74(6):970--974, 2012.

\bibitem{HBP}
H.~Markram, K.~Meier, T.~Lippert, S.~Grillner, R.~Frackowiak, S.~Dehaene,
  A.~Knoll, H.~Sompolinsky, K.~Verstreken, and J.~DeFelipe.
\newblock Introducing the human brain project.
\newblock {\em Procedia Computer Science}, 7:39--42, 2011.

\bibitem{serrano2013proposal}
T.~Serrano-Gotarredona, T.~Prodromakis, and B.~Linares-Barranco.
\newblock A proposal for hybrid memristor-{CMOS} spiking neuromorphic learning
  systems.
\newblock {\em Circuits and Systems Magazine, IEEE}, 13(2):74--88, 2013.

\bibitem{MotionParallax}
E.~C. Sobel.
\newblock The locust's use of motion parallax to measure distance.
\newblock {\em J. of Comparative Physiology A}, 167(5):579--588, 1990.

\bibitem{reichardt1956geschwindigkeitsverteilung}
H.~Reichardt.
\newblock {\"U}ber die geschwindigkeitsverteilung in einer geradlinigen
  turbulenten couettestr{\"o}mung.
\newblock {\em ZAMM-J. of Applied Mathematics and Mechanics/Zeitschrift f{\"u}r
  Angewandte Mathematik und Mechanik}, 36(S1):S26--S29, 1956.

\bibitem{microsaccades}
S.~Martinez-Conde, S.~L. Macknik, and D.~H. Hubel.
\newblock The role of fixational eye movements in visual perception.
\newblock {\em Nature Reviews Neuroscience}, 5(3):229--240, 2004.

\bibitem{NeuromorphicSytems}
C.~Mead.
\newblock Neuromorphic electronic systems.
\newblock {\em Proc. of the IEEE}, 78(10):1629--1636, 1990.

\end{thebibliography}
